\documentclass[journal]{IEEEtran}
\usepackage{amsmath,amsfonts}
\usepackage{algorithm}
\usepackage{array}
\usepackage[caption=false,font=normalsize,labelfont=sf,textfont=sf]{subfig}
\usepackage{textcomp}
\usepackage{stfloats}
\usepackage{url}
\usepackage{verbatim}
\usepackage{graphicx}
\usepackage{cite}
\usepackage{microtype}
\usepackage{multirow}
\usepackage{graphicx}
\usepackage{colortbl}
\usepackage{xcolor}
\usepackage{pifont}
\usepackage{cite}
\usepackage{cleveref}
\usepackage{amsmath}
\usepackage{ctable}
\newcommand{\eg}{\textit{e}.\textit{g}.}
\definecolor{mygray}{gray}{0.95}
\crefname{figure}{Fig.}{Figs.}
\crefname{table}{Tab.}{Tabs.}

\begin{document}

\title{Event-Aware Instructed Assistant for Referring Video Segmentation}

\author{Jinyu Liu, Henghui Ding, Shuting He, Yu-Gang Jiang\IEEEmembership{~IEEE Fellow}
\thanks{Jinyu Liu and Henghui Ding are with the Institute of Big Data, College of Computer Science and Artificial Intelligence, Fudan University, Shanghai, China (e-mail: liujy24@m.fudan.edu.cn, henghui.ding@gmail.com).}
\thanks{Yu-Gang Jiang is the Institute of Trustworthy Embodied AI, Fudan University, Shanghai, China (e-mail: ygj@fudan.edu.cn).}
\thanks{Shuting He is with Shanghai University of Finance and Economics, China (e-mail: shuting.he@sufe.edu.cn).}
\thanks{Corresponding authors: Henghui Ding and Yu-Gang Jiang}
\thanks{This project was supported by the National Natural Science Foundation of China (NSFC) under Grant No. 62472104. Henghui Ding was supported by Xiaomi Young Scholars Program.}
}

\markboth{IEEE TRANSACTIONS ON IMAGE PROCESSING }%
{Shell \MakeLowercase{\textit{et al.}}: A Sample Article Using IEEEtran.cls for IEEE Journals}


\maketitle

\begin{abstract}
Existing referring video segmentation methods often treat a video as a single event consisting of multiple images, overlooking the fact that a video typically contains multiple distinct events. Under such a mechanism, the model needs to directly understand all the complex content in the video and text, which can easily lead to confusion and hallucinations. To address this issue, we propose to decompose a video to a set of simple events by learnable Event Query, and understand complex video content in an event-by-event, easy-to-understand manner. This is based on the observation that natural language expressions often divide a video into distinct, text-related segments, each representing a separate event within a compound event. We introduce \textbf{EVIS}, an \textbf{E}vent-Aware \textbf{V}ideo \textbf{I}nstructed \textbf{S}egmentation Assistant, which utilizes text-guided Event Queries to partition a video into simple events, extracting event-aware visual-text features to achieve a hierarchical understanding of the video. Additionally, we propose Object-Pixel-Hybrid Learning, which enables the MLLMs to track targets in long-term videos by integrating fine-grained pixel features with prior object queries. Extensive experimental results on 5 public benchmarks demonstrate EVIS’s strong performance in addressing the referring video segmentation task.
\end{abstract}


\section{Introduction}
\label{sec:introduction}

\IEEEPARstart{R}{eferring} Video Object Segmentation (RVOS)~\cite{mevis,dshmp,refer-ytb-vos,ref-dacvis17} is an emerging and challenging task focused on segmenting target objects according to the given natural language description throughout a video. It has a wide range of applications in the real world, such as embodied perception and video editing. Recent datasets like MeViS~\cite{mevis} emphasize the temporal motion properties of videos, which are crucial in this field.
With the rapid development of multi-modal Large Language Models (MLLMs)~\cite{internvl,minigpt-4}, LLM-based methods like LISA~\cite{lisa} and GSVA~\cite{gsva} are introduced to referring image segmentation, showcasing the ability of MLLMs to localize and segment objects through advanced reasoning and understanding. Following LISA~\cite{lisa}, VISA~\cite{visa} and VideoLISA~\cite{videolisa} extend the use of MLLMs to the referring video domain.

\begin{figure}[htp]
\centering
\includegraphics[width=0.48\textwidth]{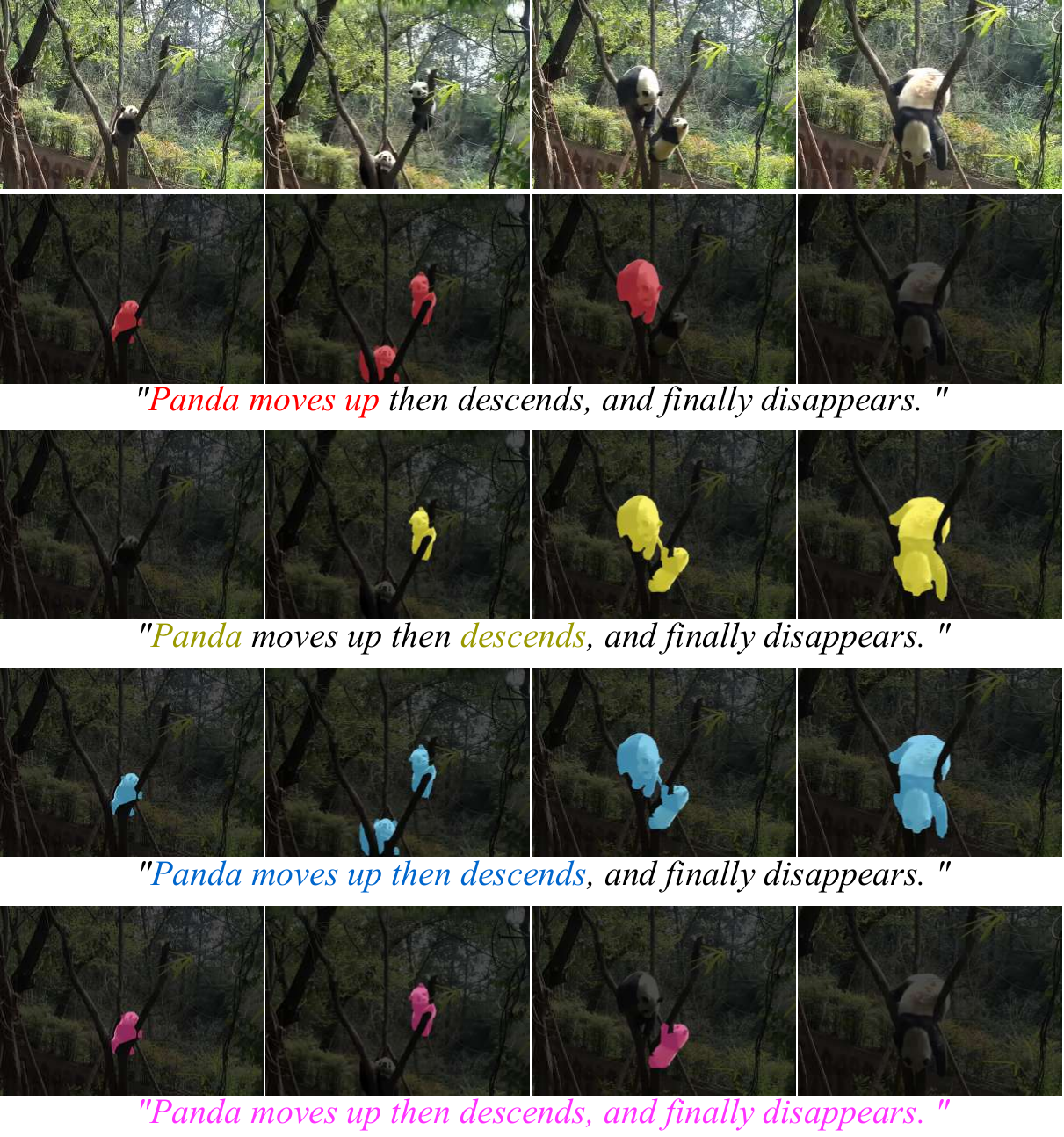}
\caption{Illustration of the Event Taxonomy~\cite{event}. When focusing on different expression parts, trajectories of related objects in the video are various. Any length of trajectory for each object can be abstracted as a simple event. These simple events make up the compound event throughout the video.}
\label{figure1}
\end{figure}

However, existing approaches~\cite{visa,videolisa,mevis,referformer,mttr,dshmp,losh} often treat a video as a single event composed of multiple images, overlooking the fact that videos with temporal motion properties~\cite{mevis} typically contain multiple distinct events. Under such a mechanism, the model needs to directly understand all the complex content in the video and text, which can easily lead to confusion and hallucinations, especially in MLLM-based methods. Even for humans, understanding complex scenarios often involves breaking down the whole into its parts and then integrating these parts into a cohesive understanding~\cite{koffka2013principles,atkinson1968human}. When processing complex information, such as videos, the human cognitive system decomposes the content into manageable segments, gradually assembling them to form an understanding of the complete scenario. Shipley et al.~\cite{event} propose the Event Taxonomy, hypothesizing that a compound event consists of the combined occurrence of two or more simple events, which occur when objects change or interact. Inspired by the Event Taxonomy~\cite{event}, we propose to decompose the video to multiple simple events by a group of learnable Event Queries and understand complex video scenarios in an event-by-event manner. We have observed that natural language expressions often segment a video into distinct, text-related parts, each representing a separate event within a compound structure. 
As shown in \cref{figure1}, when considering the partial phrase \textit{“panda moves up”}, two object trajectories from the earlier section align with it. Similarly, the phrase \textit{“panda descends”} corresponds to the latter part of this case. However, for \textit{“panda moves up then descends”}, all objects meet the criteria: the blue segments in the 3rd row of \cref{figure1}. This illustrates the importance of distinguishing between simple events within a video sequence for accurate interpretation of the full video content in relation to the given linguistic description.

\begin{figure*}[htp]
\centering
\includegraphics[width=0.98\textwidth]{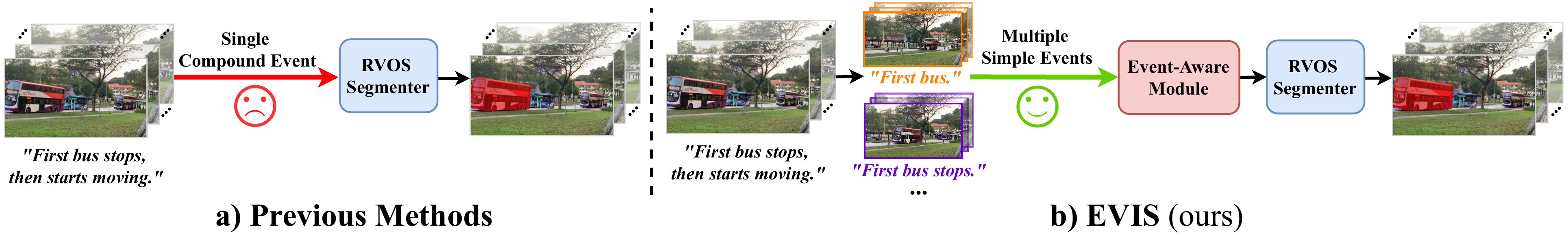}
\caption{Comparison of previous methods and ours. a) Previous methods often struggle to directly understand the complex content within a single video. b) \textbf{EVIS} (ours) innovatively shifts from single compound-event video paradigm to hierarchical simple-event analysis.}
\label{figure2}
\end{figure*}

With Event Query to decompose the given video into simple events, we introduce \textbf{EVIS}, \textbf{E}vent-Aware \textbf{V}ideo \textbf{I}nstructed \textbf{S}egmentation Assistant, to address the challenging referring video segmentation. The proposed EVIS leverages text-guided Event Queries to partition a video into distinct simple events, extracting event-aware visual-text features to enable a hierarchical understanding of the video content. The event-centric design distinguishes EVIS from previous RVOS methods, as shown in~\cref{figure2}. To capture events across frames throughout the video, EVIS incorporates an advanced Event-Aware Frame Merging Module (EAFM), which uses text-guided Event Queries to merge objects from multiple frames into a set of specific events. Additionally, Event-Intra Attention and Event-Inter Attention are integrated to the EAFM. The Event-Intra Attention enhances the model's ability to capture the short-term spatial-temporal dynamics in each event, while the Event-Inter Attention supports long-term learning across multiple events. These blocks not only facilitate fine-grained learning within events but also maintain a coherent understanding throughout the entire video, ensuring robust interactive alignment between video content and textual expressions.
However, pixel features tend to be highly redundant, while object features, though fewer in number, capture more abstract and meaningful information. To tackle this issue, we introduce Object-Pixel-Hybrid Learning, which encourages the simultaneous learning of prior object queries and pixel-level features by a single \texttt{[SEG]} token for each video, achieving a multi-level feature interaction. This learning strategy guarantees long-term object tracking while maintaining computational efficiency. Overall, this simple-to-compound learning procedure allows EVIS to preserve object coherence in complex scenes.

In summary, our main contributions are as follows:
\begin{itemize}
\item We propose \textbf{EVIS}, an \textbf{E}vent-Aware \textbf{V}ideo \textbf{I}nstructed \textbf{S}egmentation Assistant that leverages the Event Query and EAFM module for hierarchical multi-modal understanding in referring video segmentation.
\vspace{1mm}
\item We design Event Query to decompose a video to a series of simple events, promoting the model to understand complex video scenarios in an event-by-event, easy-to-understand manner.
\vspace{1mm}
\item To capture events, we develop an Event-Aware Frame Merging Module (EAFM) combined with Event-Intra Attention and Event-Inter Attention to achieve comprehensive video-text alignment.
\vspace{1mm}
\item We demonstrate impressive performance across five referring video segmentation datasets, notably achieving a 46.8\% \( \mathcal{J} \)\&\( \mathcal{F} \) on the challenging MeViS dataset, validating the effectiveness of the proposed approach.
\end{itemize}

\section{Related Work}
\label{sec:related_work}

\begin{figure*}[t]
\centering
\includegraphics[width=0.98\textwidth]{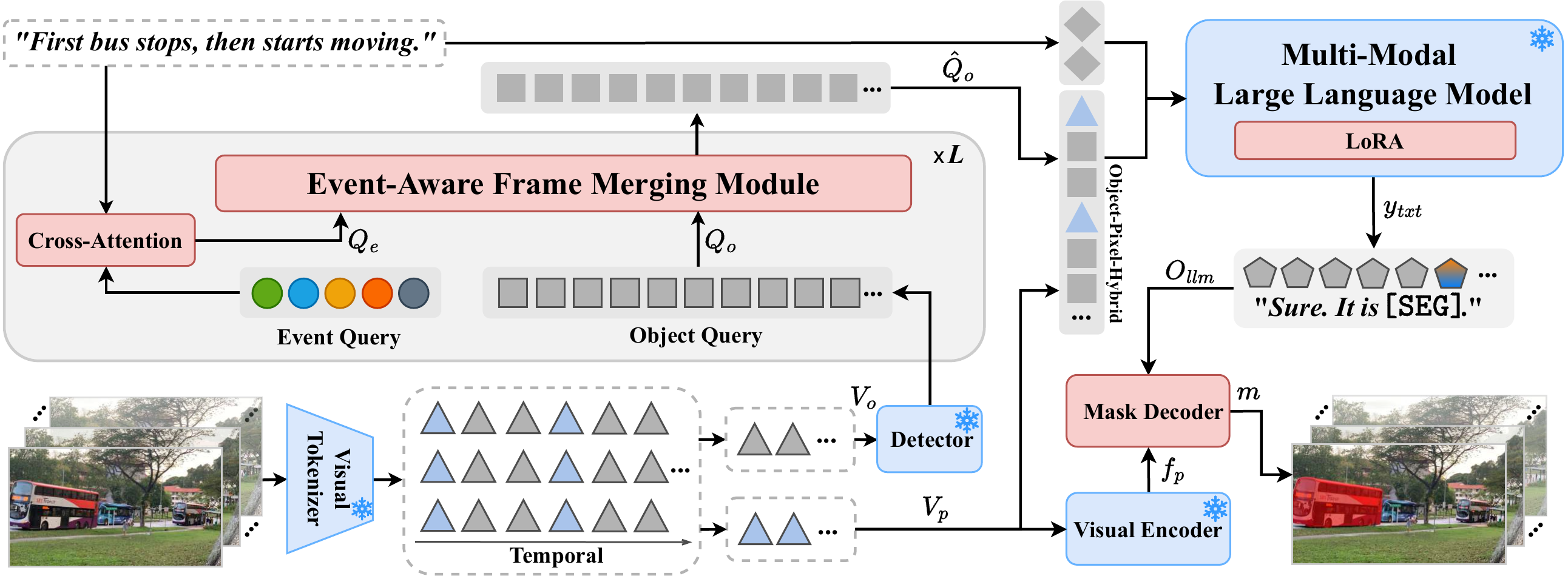}
\caption{Overview of the proposed \textbf{E}vent-Aware \textbf{V}ideo \textbf{I}nstructed \textbf{S}egmentation Assistant (\textbf{EVIS}). EVIS employs the event query and Event-Aware Frame Merging Module (EAFM) to learn hierarchical video features in an event-by-event manner. First, we decouple the visual tokens to pixel tokens $V_{p}$ and object tokens $V_{o}$, {where tokens
are split along the temporal dimension}. Using object queries $Q_{o}$ generated by the detector, EAFM module is applied to comprehend various objects, guided by the event query. Notably, the event query cross-attends with the text embeddings at the beginning of each EAFM block, ensuring full interactions between video and text and preventing text information from being lost. Finally, the pixel tokens $V_{p}$ and updated event-level object queries $\hat{Q}_{o}$ are fed into the MLLMs to generate the \texttt{[SEG]} token and predict the final mask, ‌facilitating comprehensive video understanding.
}
\label{figure3}
\end{figure*}

\textbf{Referring Video Segmentation} focuses on segmenting targets within a video according to the given expression. The field has been significantly accelerated by several benchmark datasets, \eg, A2D-Sentences~\cite{a2d-sentence}, Ref-DAVIS17~\cite{ref-dacvis17}, Ref-YouTube-VOS~\cite{refer-ytb-vos}, and MeViS~\cite{mevis}. Some previous methods directly extend referring image segmentation to video scenes. For example, Khoreva et al.\cite{ref-dacvis17} incorporates the referring image segmentation methods MAttNet\cite{mattnet} to segment each frames and use post-processing to maintain temporal coherence. Other studies, such as ReferFormer~\cite{referformer} and MTTR~\cite{mttr}, introduce a DETR-like architecture for referring video object segmentation, streamlining the segmentation process and achieving notable results. More recently, MeViS~\cite{mevis} dataset, based on the complex video object segmentation dataset MOSE~\cite{mose}, is developed to emphasize the importance of motion expressions and address current methods’ limitations in understanding motion information. Some previous methods, such as VISA~\cite{visa} and VideoLISA~\cite{videolisa}, leverage MLLMs for this task: VISA~\cite{visa} extends the referring image segmentation task to the video domain by propagating the first frame to the rest video frames through an object tracking method~\cite{xmem}, while VideoLISA~\cite{videolisa} proposes a sparse dense sampling strategy to reduce the number of tokens, preserving temporal dynamics by applying global average pooling on sparse frames to reduce them to a lower resolution. {Recently, Glus~\cite{glus} adopts Global and Local consistency, a set of sparse context frames provides global information, while a stream of continuous query frames conducts local object tracking. Villa~\cite{glus} proposes Key Segment Extractor to select the most query-relevant segments, and uses Context Synthesizer to aggregate text-related visual cues for addressing long durations, multiple objects, rapid motion, and heavy occlusions. Veason-R1~\cite{Veason-R1} explores the effect of Group Relative Policy Optimization (GRPO)~\cite{grpo} for video reasoning segmentation.}
In this study, we focus on enhancing the understanding of temporal motion cues in an event-by-event, easy-to-understand manner.

\textbf{Referring Image Segmentation}
involves segmenting targets within images based on text descriptions. Methods are mainly categorized into one-stage~\cite{vlt,Hu,vlt+tc,LAVT} that perform end-to-end predictions and two-stage~\cite{mattnet,liu2022instance,Chen_lang2seg_2019} that decompose the task into instance segmentation and text-instance matching. Hu et al.~\cite{Hu} exemplify the one-stage paradigm by integration of visual and linguistic features for mask prediction. MAttNet~\cite{mattnet} demonstrates the two-stage way by initially employing instance segmentation, followed by text-guided object selection. The emergence of Transformer \cite{transformer} has catalyzed significant advancement in referring image segmentation. Ding et al.~\cite{vlt} pioneers the application of Transformers in this domain through their Vision-Language Transformer (VLT). This breakthrough has precipitated the development of numerous Transformer-based methods. Notable examples include LISA~\cite{lisa} and GSVA~\cite{gsva}. LISA~\cite{lisa} introduces a specialized \texttt{[SEG]} token to interface with segmentation mask decoders such as SAM~\cite{segment-anything}, enabling MLLMs to generate precise masks. Building upon this foundation, GSVA~\cite{gsva} extends the functionality of \texttt{[SEG]} token while introducing \texttt{[REJ]} token to support generalized referring expression segmentation (GRES)~\cite{GRES}. {VRS-HQ~\cite{devil} proposes Temporal Dynamic Aggregation and Token-driven Keyframe Selection modules achieving strong performance.}

\textbf{Multi-Modal Large Language Model}.
The rapid development of large language models (LLMs) motivates research into extending their capabilities to the visual domain, breeding the multi-modal large language models (MLLMs)~\cite{llava,blip2,minigpt-4,internvl,dualfocus,qwen-vl,llava1.5,bliva,chatunivi}. Previous work like LLaVA~\cite{llava}, BLIP-2~\cite{blip2}, MiniGPT-4~\cite{minigpt-4} and Qwen-VL~\cite{qwen-vl}, empowering the MLLMs with ability to handle complex visual tasks. In video scenarios, learning temporal information becomes crucial. A straightforward way is to concatenate tokens from multiple frames, , though this is constrained by computational limits. To address this issue, some works~\cite{chatunivi, imagepoints, dynamic-vit, testa} employ token merging or pooling strategies, which reduce token numbers at the cost of detail loss. Other works such as BLIP-2~\cite{blip2} use Q-Former architecture to extract abstract features, retaining rich information while reducing token count. Recent studies~\cite{internvl, dualfocus, qwen-vl, llava1.5, bliva} have further integrated region-level image grounding and pixel-level understanding into MLLMs. For example, InternVL~\cite{internvl} aligns the representation of the scaled-up vision encoder with LLMs and achieves surprising performance on image grounding task.

\section{Method}
\label{sec:method}

{\subsection{Preliminaries}}
{
$V$: Visual tokens of the input video. 
$V_p$: Pixel tokens, input for the visual encoder.
$V_o$: Object tokens, input for the detector.
$Q_o$: Object queries.
$\hat{Q}_o$: Object queries updated by the EAFM module.
$Q_{o'}$: Object queries from all events. 
$q_o$: Object queries from one frame. 
$F_s$: Sentence embedding. 
$Q_e$: Event queries. 
$Q_g$: Global object queries. 
$k$: Number of top event queries selected in the Frame Merging Block. 
$\mathcal{A}$: Assign attention in the Frame Merging Block, which is not differentiable. 
$\mathcal{\hat{A}}$: Assign attention, preserving gradient of $\mathcal{A}$ and is differentiable. 
$\mathcal{\Tilde{A}}$: Attention between $\hat{E}$ and $\hat{Q}_o$. 
$W_o$: Learnable transformation matrices for object queries in the Gumbel Softmax operation.
$W_e$: Learnable transformation matrices for event queries in the Gumbel Softmax operation.
$W_{o'}$: Learnable transformation matrices for object queries in the process of global object queries generation. 
$W_{e'}$: Learnable transformation matrices for event queries in the process of global object queries generation. 
$E$: Object queries in single event. 
$\hat{E}$: Object queries in single event updated by the Event-Intra Attention. 
$L$: Number of stacking layers of the EAFM module. 
$f_p$: Pixel features generated by the ViT of MLLM. 
$N_l$: Number of frames in single video.
$N_o$: Number of object queries in single frame. 
$N_{o'}$: Number of frames in single event. 
$N_h$: Number of event queries.
$i$: Index of object query in single frame. 
$l$: Index of frame in single video. 
$h$: Index of event queries in single video. 
$t$: Index of set of object queries in single event. 
$U$: Hybrid unit. 
$T_o$: Number of object queries in a hybrid unit. 
$T_u$: Number of pixel features in a hybrid unit. 
$m$: Ground truth segmentation mask. 
$\hat{m}$: Predicted segmentation mask by the model. 
$\mathcal{M}$: Event attention mask for object queries in the Event-Intra Attention. 
$y_t$: Ground truth text answer. 
$\hat{y}_t$: Text answer generated by the LLM. 
$O_{llm}$: \texttt{[SEG]} token embedding of the outputs generated by the LLM. 
$\lambda_{m}$: Weight of mask loss $\mathcal{L}_m$. 
$\lambda_{t}$: Weight of text generation loss $\mathcal{L}_t$. 
$\lambda_{bce}$: Weight of BCE loss. 
$\lambda_{dice}$: Weight of DICE loss. 
}

An overview of the proposed EVIS approach is shown in~\cref{figure3}. First, we decouple the visual tokens $V$ into pixel tokens $V_{p}$ and object tokens $V_{o}$ using a visual tokenizer. {To ensure dimensional consistency for input into the MLLM, we use a projection layer to adjust the feature dimension of $V_{o}$.} Mask2Former~\cite{mask2former} is used as the detector to extract object queries $Q_{o}$ of potential candidate objects. Text features are extracted by RoBERTa~\cite{roberta}. The event query $Q_{e}$ embeds the textual features by cross-attending to the sentence embedding $F_{s}$ before being inputted into the Event-Aware Frame Merging Module (EAFM). EAFM is applied to the object queries $Q_{o}$ to progressively gather spatial-temporal information, generating event-level object queries $\hat{Q}_{o}$ under the guidance of the event query $Q_{e}$. EAFM module is stacking with $L$ layers. Next, text tokens $V_{t}$, pixel tokens $V_{p}$, and object queries $\hat{Q}_{o}$ are fed into the MLLMs to produce the \texttt{[SEG]} token. Finally, the last-layer embedding of the \texttt{[SEG]} token is decoded into a segmentation mask via the mask decoder.

\subsection{Easy-to-Understand: Event Query}
Existing methods~\cite{visa, videolisa, mevis, referformer, mttr, dshmp, losh} for referring video segmentation often treat a video as a single event. These methods are not specifically designed for the complex and varied scenes within videos and primarily rely on referring image segmentation paradigms to perform frame-by-frame target object segmentation.

We argue that the model can better understand complex, compound events in a video if it first comprehends the simple events within them. Shipley et al.~\cite{event} propose the Event Taxonomy, hypothesizing that a compound event consists of the combined occurrence of two or more simple events, occurring when objects change or interact. Inspired by this taxonomy, we design the learnable Event Query to decompose the video into simple events, based on the observation that natural language expressions often divide a video into distinct, text-related segments, each representing a separate event within a compound event. {Notably, our method differs fundamentally from object trajectory re-grouping in both representation and supervision. Instead of hard-assigning objects to disjoint temporal segments, we model events as dynamic, semantic attention aggregations over object queries, which allows the same object to simultaneously participate in multiple events.} The Event Query encourages the model to understand complex video scenarios on an event-by-event, easy-to-understand basis.

\begin{figure}[t]
\centering
\includegraphics[width=0.46\textwidth]{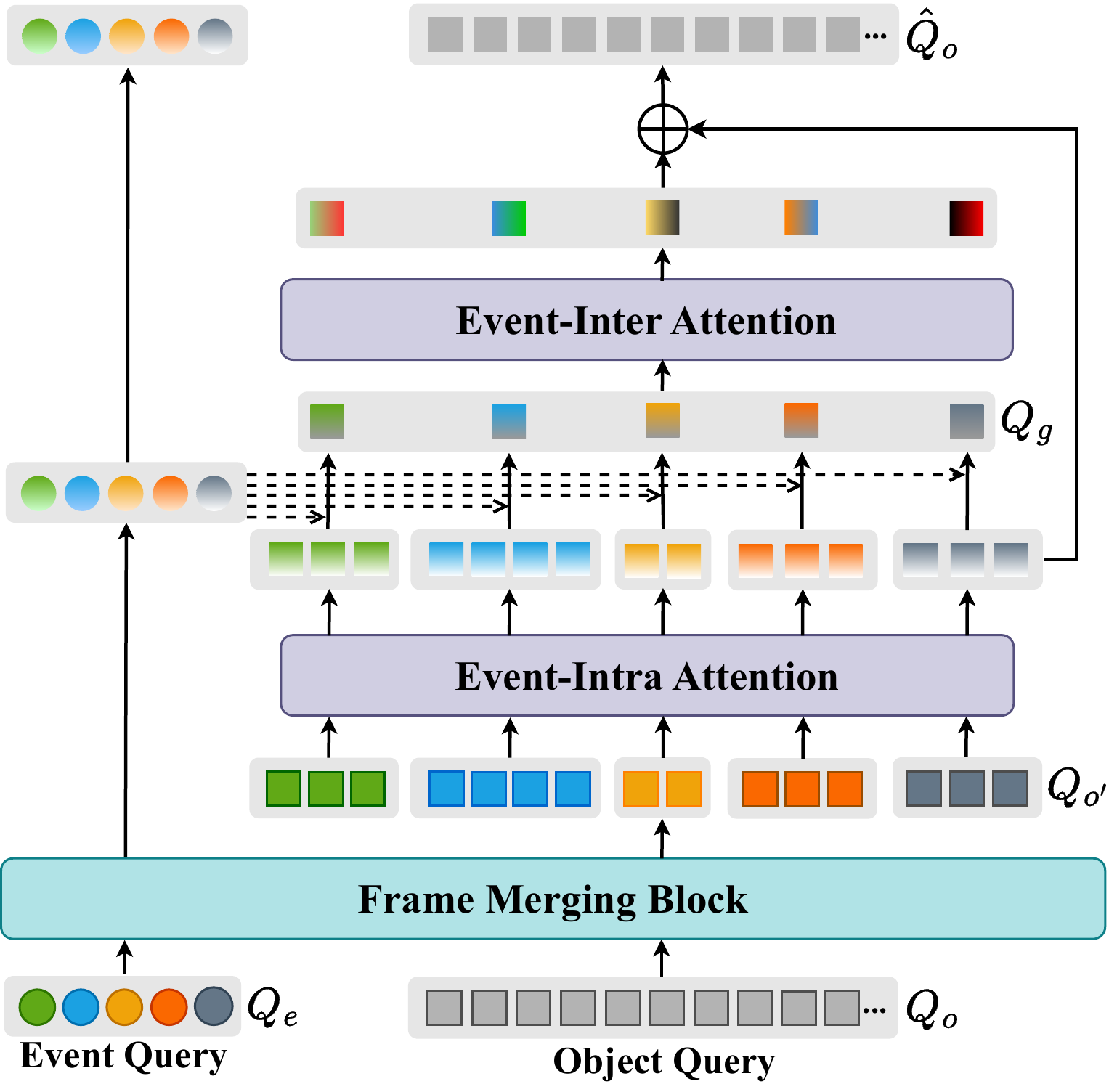}
\caption{\textbf{Event-Aware Frame Merging Module} (\textbf{EAFM}). The EAFM module effectively comprehends various objects in an event-by-event manner, guiding the MLLMs to capture event-intra and event-inter information in a video.}
\label{figure4}
\end{figure}

\subsection{Event-Aware Frame Merging Module}
{EAFM aggregates frame-wise object queries into temporal events and models both intra-event dynamics and inter-event relationships}. \cref{figure4} shows the proposed Event-Aware Frame Merging Module. First, the frame merging block group object queries into distinct simple events. Subsequently, an event-intra attention mechanism is applied to capture fine-grained spatial-temporal information within each event. Following this, cross-attention is performed between the event query and the refined event-intra object queries to produce a global query for each event. These global object queries then undergo event-inter attention, allowing for the extraction of long-term object trajectory features, which generate the event-inter object queries. Finally, the resulting sequence of event-level object queries is computed as a weighted sum of event-intra and event-inter object queries.

\begin{figure}[t]
\centering
\includegraphics[width=0.46\textwidth]{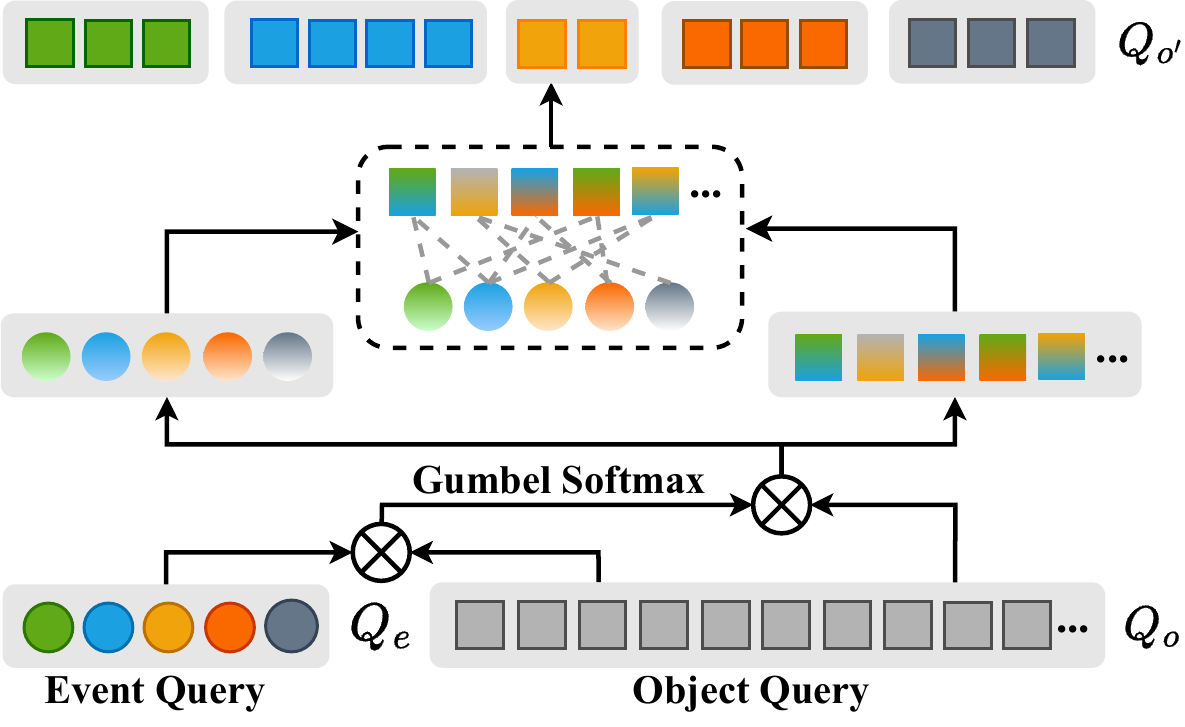}
\caption{\textbf{Frame Merging Block}. We compute the event to assign an object query to by selecting the top-$k$ Event Queries.}
\label{figure5}
\end{figure}

\noindent\textbf{Frame Merging Block} {groups object queries into distinct events based on embedding similarity using differentiable top-$k$ assignment}. Rather than forwarding image tokens from all $N_l$ frames, we use object queries generated by the detector, Mask2Former~\cite{mask2former}, to ensure efficient interaction. As shown in~\cref{figure5}, the Frame Merging Block receives the learned event query and object query as inputs. It merges all object queries associated with the same event query into a new sequence of object queries, based on their similarity within the embedding space.

Formally, suppose there are $N_{h}$ events in a video, each indexed by $h$ and with a set of learnable event queries $\{Q_e^{h}\}_{h=1}^{N_h}$. We denote the object queries from all the frames as $\{Q_o^{l}\}_{l=1}^{N_l}$, the $l$-th frame's object queries $Q_o^l=\{{q}_o^i\}_{i=1}^{N_o}$, where $N_l$ is the frame number and $N_o$ is the object query number in a single frame. We simplify $\{Q_e^{h}\}_{h=1}^{N_h}$ to $\{Q_e^{h}\}$ and similarly$\{Q_o^{l}\}_{l=1}^{N_l}$ to $\{Q_o^{l}\}$. The association between the event queries $\{Q_e^{h}\}$ and object queries $\{Q_o^{l}\}$ is established through a similarity matrix $\mathcal{A}$ inspired by~\cite{groupvit}, computed using the \text{Gumbel Softmax}~\cite{gumble-softmax} operation:
\begin{equation}
    \mathcal{A}_{m, n} = \frac{\exp(W_e{Q}_e^{m} {\cdot} W_o{Q}_o^{n} + \gamma_m)} {\sum_{h=1}^{N_h} \exp(W_e{Q}_e^{h} {\cdot} W_o{Q}_o^{n} + \gamma_h)},
\end{equation}
where the $W_e$ and $W_o$ serve as learnable transformation matrices that project the event and object queries into a shared embedding space, while the stochastic variables $\{\gamma\}$ are drawn independently from the \text{Gumbel(0, 1)} distribution. We compute the event to assign an object query to by selecting the \text{top-}$k$ events. Since the \text{top-}$k$ assignment operation is not differentiable, to address this, we use a straightforward trick in~\cite{trick} to compute as:
\begin{equation}
\hat{\mathcal{A}} = \text{top-}k({\mathcal{A}}) + \mathcal{A} - \text{sg}(\mathcal{A}),
\end{equation}
where $\text{sg}$ denotes the stop gradient operator. This formulation enables $\hat{\mathcal{A}}$ to maintain the \text{top-}$k$ discrete event assignments while preserving gradient of $\mathcal{A}$, ensuring the Frame Merging Block remains end-to-end optimization.

\noindent\textbf{Event-Intra Attention}. To capture the short-term spatial-temporal dynamics within each event {via masked self-attention}, we propose the Event-Intra Attention block. This block applies a mask $\mathcal{M}$ to each event, ensuring that only the information relevant to the specific event is considered, thereby isolating it from external influences. We denote the object queries from all events as $Q_{o'} = \{E_{h}\}_{h=1}^{N_h}$. Assuming that the $h$-th event $E_h=\{Q_{o'}^{t}\}_{t=l'}^{l' + N_{o'}}$ begins at the $l'$-th set of object queries, where $N_{o'}$ is the number of object query sets in the current event. Event-intra Attention is calculated as:
\begin{equation}
\mathcal{M}_h[m,n] = 
\begin{cases}
1, & \text{if } ({Q}_{o'}^{m} \in {E}_h)  \land  ({Q}_{o'}^{n} \in {E}_h) \\
-\infty, & \text{otherwise}
\end{cases}
,
\end{equation}

\begin{equation}\label{eq:intra-attention}
   \hat{E}_h = \text{softmax}\left(\mathcal{M}_h \odot \frac{{E}_h {Q}_{o'}^{T}}{\sqrt{C}}\right){Q}_{o'},
\end{equation}
where $C$ is the number of channels, $\mathcal{M}_h$ is the event attention mask for the $h$-th event's object queries ${E}_h$, and $\odot$ denotes element-wise multiplication.

\noindent\textbf{Global Object Query} {aggregates intra-event object queries into a compact event-level representation}. Based on the object queries updated by the Event-Intra Attention, we further abstract the object queries into a global object query $Q_{g}^{h}$ for the $h$-th event. This process is guided by the updated event query through cross-attention and a weighted sum operation:
\begin{equation}
    Q_{g}^{h} = W_{e'}\frac{\sum_{t={l'}}^{{l'}+N_{o'}}\mathcal{\Tilde{A}}  W_{o'} {E}_{h} } { \sum_{t={l'}}^{{l'}+N_{o'}}\Tilde{\mathcal{A}}}
    ,
\end{equation}
where $W_{e'}$ and $W_{o'}$ are learned weights for the event and object queries to obtain merged features. $\Tilde{\mathcal{A}}$ is the attention matrix between the updated event and object queries.

\noindent\textbf{Event-Inter Attention} {models long-range temporal dependencies and object trajectories across different events}. To capture long-term event trajectory features across all events, we further design an Event-Inter Attention block. The global object query generated by Event-Intra Attention is used as input to Event-Inter Attention, which is computed similarly to Event-Intra Attention but without applying the event mask.

\noindent\textbf{Event Integration} {fuses intra-event and inter-event features and maps enriched representations back to original positions}. Through the propagation of the Event-Intra Attention and Event-Inter Attention, we capture both internal and external event information. In order to integrate the features across various temporal dimensions, we conduct event integration between the event-intra object queries and event-inter global object queries. During each block calculation, we record the position of the initial object query at each step and average the intra-event and inter-event queries, mapping the updated results back to the original input locations.

\subsection{Object-Pixel-Hybrid Learning}
While models are becoming increasingly capable with advancements in MLLMs, the number of model parameters is also growing. VideoLISA~\cite{videolisa} employs a sparse dense sampling strategy to reduce the number of visual tokens, applying global average pooling in sparse frames to lower their resolution. However, directly averaging high-dimensional features can cause information distortion and result in a loss of important details.

To address this challenge, we propose Object-Pixel-Hybrid Learning, which enables MLLMs to track targets in long-term videos by integrating fine-grained pixel features with contextual object queries. In a single frame, pixel features tend to be highly redundant, while object features, though fewer in number, capture more abstract and meaningful information. Additionally, object features can guide the language model in filtering out redundant information from the pixel features. We define a hybrid unit composed of a single pixel feature and multiple object queries, allowing for multi-level feature representation and interaction. Formally, we concatenate the pixel features $f_p = \text{ViT}(V_{p})$ with event-level object queries $\hat{Q}_o$, where ViT is the vision encoder of the MLLMs, and then interlace them as input into the LLMs:
\begin{equation}
    U = \text{Concat}[f_{p}^{l}, \hat{Q}_o^{l + 1}, ..., \hat{Q}_o^{l + T_{o}}], 
\end{equation}
\begin{equation}
        O_{llm} = {
        \text{LLMs}}(\text{Concat}[U_{1}, ..., U_{T_{u}}]),
\end{equation}
where $T_{o}$ is the number of object queries in a hybrid unit, $U$ denotes the hybrid unit consisting of one single pixel feature $f_p$ and $T_{o}$ object queries, $\text{Concat}[~\!,]$ is concatenation. $T_{u}$ is the number of hybrid units, and $O_{llm}$ refers to the outputs of the LLMs. With this hybrid input, we aim to encourage the model to learn features of varying dimensions, while mitigating the negative influence of redundant pixel features.

\subsection{Training Objective}
With Object-Pixel-Hybrid Learning, an apparent issue is the disparity between pixel and target features. To address this, we select an equal number of pixel and object features to generate masks for the loss calculation. 
Building upon the foundational work LISA~\cite{lisa}, the overall training objective $\mathcal{L}$ is formulated as a weighted combination of the standard text generation loss $\mathcal{L}_{t}$ and segmentation mask loss $\mathcal{L}_{m}$:
\begin{equation}
    \mathcal{L} = \lambda_{t} \mathcal{L}_{t} + \lambda_{m} \mathcal{L}_{m},
\end{equation}
where $\mathcal{L}_{t}$ is the auto-regressive cross-entropy (CE) loss for text generation, while $\mathcal{L}_{m}$ combines per-pixel binary cross-entropy (BCE) loss and DICE loss~\cite{dice}, weighted by their respective coefficients $\lambda_{bce}$ and $\lambda_{dice}$. Given the ground truth $(\hat{y}_{txt}, \hat{m})$ and model predictions $(y_{txt}, m)$, $\mathcal{L}_{txt}$ and $\mathcal{L}_{m}$ are defined as:
\begin{equation}
    \mathcal{L}_{txt} = \text{CE}(\hat{y}_{txt}, y_{txt}),
\end{equation}
\begin{equation}
    \mathcal{L}_{m} = \lambda_{bce}\text{BCE}(\hat{m}, m) + \lambda_{dice}\text{DICE}(\hat{m}, m),
\end{equation}
where $\hat{y}_{txt}$, $y_{txt}$ correspond to textual sequences and $\hat{m}$, $m$ denote binary segmentation masks.

\subsection{Re-implemented EVIS without LLM.}
\begin{figure}[h]
\centering
\includegraphics[width=0.46\textwidth]{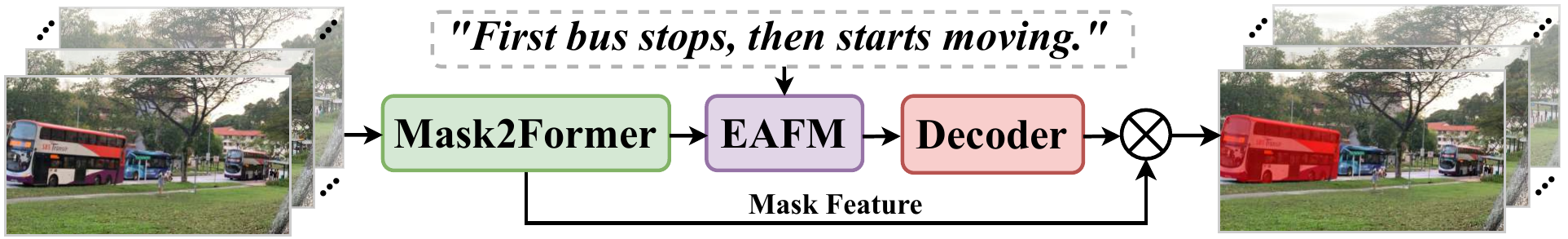}
\caption{Model architecture of the re-implemented EVIS by removing LLM.}
\label{figure7}
\end{figure}

Since the event-centric design is independent of the model's reliance on LLMs, a natural question arises: can the EAFM module operate without LLM? To investigate this possibility, we build a simple baseline model by removing the LLM and SAM components to evaluate the capability of EAFM, as shown in~\cref{figure7}. For experiments, we follow~\cite{dshmp}. ~\cref{table:sota_refer} additionally compares the re-implemented EVIS with SOTA methods without LLMs.

\section{Experiments}
\label{sec:experiments}
\subsection{Datasets and Evaluation Metrics}
\noindent\textbf{Datasets}.
The proposed EVIS approach is trained on a variety of segmentation datasets. For image datasets, we follow LISA~\cite{lisa}. For video datasets, we follow~\cite{videolisa} to optimize the EAFM module for video-text dynamics. We evaluate across five RVOS benchmarks: MeViS~\cite{mevis}, Ref-YouTube-VOS~\cite{refer-ytb-vos}, Ref-DAVIS17~\cite{ref-dacvis17}, A2D-Sentences~\cite{a2d-sentence}, and JHMDB-Sentences~\cite{jhmdb}. MeViS is a recently established benchmark focusing on motion analysis and contains 2,006 videos with 28K annotations. Ref-YouTube-VOS is the most extensive RVOS dataset, containing 3,978 videos and 13K text annotations. Ref-DAVIS17 adds text descriptions to DAVIS17~\cite{davis17} dataset. A2D-Sentences includes 3.7K videos with 6.6K action-specific annotations, and JHMDB-Sentences has 928 videos across 21 action categories.

\noindent\textbf{Evaluation Metrics}. We use region similarity $\mathcal{J}$, contour accuracy $\mathcal{F}$, and their average $\mathcal{J}$\&$\mathcal{F}$. For A2D-Sentences and JHMDB-Sentences, we use mAP, oIoU, and mIoU.

\subsection{Implementation Details}
We implement our model using InternVL2~\cite{internvl}, a multimodal large language model built upon InternViT~\cite{internvit} and Qwen2~\cite{qwen2} with 1B parameters. The visual encoder and mask decoder are derived from SAM~\cite{segment-anything}. We perform joint training with both image and video datasets. In the pre-processing stage, we insert the original category name or referring expression from the dataset into a template. For example: \textit{“USER: \texttt{<VIDEO>} Can you segment \texttt{\{description\}} in this video? ASSISTANT: Sure, it is} \texttt{[SEG]}.”, where \texttt{\{description\}} serves as a placeholder for the specific text to fill. For the video data, we configure $T_{u}=4$ and $T_{o}=8$. For the EAFM module, we set $L=3$, $N_h=6$ and $k=2$, respectively. In the case of image data, we replicate the images to create pseudo video sequences. We apply the AdamW~\cite{adamw} optimizer, setting the learning rate to 0.0012 and the weight decay to 0. The learning rate scheduler is WarmupDecayLR, with 100 warmup iterations. Both the weights for text generation loss $\lambda_{t}$ and mask loss $\lambda_{m}$ are set to 1.0, while the weights for BCE loss $\lambda_{bce}$ and DICE loss $\lambda_{dice}$ are set to 2.0 and 0.5, respectively. The per-device batch size is configured to 8. We use a total of 6,000 iterations to train the final model.

\begin{table}[t]
    \footnotesize
    \centering
    \setlength\tabcolsep{14pt}
    \caption{Ablation study of our proposed EVIS on MeViS dataset. EAFM and OPH represent Event-Aware Frame Merging Module and Object-Pixel-Hybrid Learning, respectively.}
    \label{ablation1}
    \begin{tabular}{cc|ccc}
    \specialrule{.1em}{.05em}{.05em}
    EAFM& OPH& $\mathcal{J}$\&$\mathcal{F}$ & $\mathcal{J}$ & $\mathcal{F}$ \\
    \hline
     \textcolor{gray}{\ding{55}} & \textcolor{gray}{\ding{55}} &39.4 &36.5 &42.3  \\
     \textcolor{gray}{\ding{55}} & \ding{51} &41.2 &38.1 &44.3  \\
     \ding{51} & \textcolor{gray}{\ding{55}} &44.9 &42.1 &47.7  \\
     \ding{51} & \ding{51} &\textbf{46.8} &\textbf{43.7} &\textbf{49.9}  \\
    \specialrule{.1em}{.05em}{.05em}
    \end{tabular}
\end{table}
\begin{table}[t]
    \centering
    \footnotesize
    \setlength\tabcolsep{11.5pt}
    \caption{Ablation study of~Event-Aware~Frame~Merging~(EAFM) module on MeViS. FM, Intra, and Inter denote Frame Merging, Event-Intra Attention, and Event-Inter Attention, respectively.}
    \label{ablation2}
    \begin{tabular}{ccc|ccc}
    \specialrule{.1em}{.05em}{.05em}
    FM& Intra& Inter& $\mathcal{J}$\&$\mathcal{F}$ & $\mathcal{J}$ & $\mathcal{F}$ \\
    \hline
     \textcolor{gray}{\ding{55}} & \textcolor{gray}{\ding{55}} & \textcolor{gray}{\ding{55}} &41.2 &38.1 &44.3  \\
     \ding{51} & \textcolor{gray}{\ding{55}} & \textcolor{gray}{\ding{55}} &45.8 &43.1 &48.5  \\
     \ding{51} & \ding{51} & \textcolor{gray}{\ding{55}} &46.1 &43.3 &48.9  \\
     \ding{51} & \textcolor{gray}{\ding{55}} & \ding{51} &46.5 &43.5 &49.5  \\
     \ding{51} & \ding{51} & \ding{51} &\textbf{46.8} &\textbf{43.7} &\textbf{49.9}  \\
    \specialrule{.1em}{.05em}{.05em}
    \end{tabular}
\end{table}
\begin{table}[t!]
    \centering
    \footnotesize
    \setlength\tabcolsep{21pt}
    \caption{Different number of event queries in the EAFM module.}
    \label{ablation3}
    \begin{tabular}{c|ccc}
    \specialrule{.1em}{.05em}{.05em}
    \multirow{1}{*}{$N_{h}$} & \multirow{1}{*}{$\mathcal{J}$\&$\mathcal{F}$} &\multirow{1}{*}{$\mathcal{J}$}  &\multirow{1}{*}{$\mathcal{F}$} \\
    \hline
         4 &46.1 &42.6 &49.6  \\
         5 &46.6 &43.2 &50.0  \\
         6 &\textbf{46.8} &\textbf{43.7} &\textbf{49.9}  \\
         7 &46.6 &43.4 &49.8  \\
    \specialrule{.1em}{.05em}{.05em}
    \end{tabular}
\end{table}

\subsection{Ablation Study}
We perform ablation studies on the challenging MeViS~\cite{mevis}.

\noindent\textbf{Module Effectiveness}.
We first conduct ablation experiments to evaluate the effectiveness of different aspects of the proposed EVIS. As shown in \cref{ablation1}, adding the Event-Aware Frame Merging Module (EAFM) yields a 5.5\% huge improvement in $\mathcal{J}$\&$\mathcal{F}$ over the baseline, which is our re-implementation based on VideoLISA~\cite{videolisa} integrated by InternVL2~\cite{internvl}. The inclusion of EAFM strengthens the model’s capacity to comprehensively capture event-level video-text features, supporting improved multi-modal understanding. Subsequently, the Object-Pixel-Hybrid Learning (OPH) is introduced to enable the MLLMs to track targets in long-term videos by integrating fine-grained pixel features with prior object queries, which contributed an additional 1.9\% gain in $\mathcal{J}$\&$\mathcal{F}$. Finally, EVIS achieve a state-of-the-art $\mathcal{J}$\&$\mathcal{F}$ of 46.8\% on the challenging MeViS dataset, underscoring the efficacy of the proposed approach.

\noindent\textbf{Ablation of EAFM Module}.
We conduct ablation experiments on the effectiveness of EAFM module. As shown in \cref{ablation2}, incorporating the Frame Merging (FM) module results in a 4.6\% improvement in $\mathcal{J}$\&$\mathcal{F}$ over the vanilla baseline. The FM module enables the model to split object queries into different simple events for afterward event-level understanding.
Next, we introduce the Event-intra Attention module to capture fine-grained short-term spatial-temporal dynamics within each event. This addition boosts performance by 0.3\% in $\mathcal{J}$\&$\mathcal{F}$, highlighting the critical role of fine-grained event-intra understanding in referring video segmentation.
Then we incorporate Event-inter Attention to further motivate the model learning object trajectory global representations between all of the events, which is essential for long-term understanding in videos. Inclusion of Event-inter Attention results in a 0.7\% gain in $\mathcal{J}$\&$\mathcal{F}$. Finally, when all components are integrated, EAFM module results in a performance improvement of 4.3\%, which demonstrates the effectiveness of the proposed method.

\noindent\textbf{Number of Event Queries $N_{h}$}. 
As mentioned above in Fig.~\ref{figure1}, any length of trajectory for each object can be abstracted as a simple event. We utilize the event query to interact with object queries from a single video generating multiple simple events. The number of events is equal to the number of event queries. When a video is partitioned into a limited number of events, each event may encapsulate an overwhelming amount of information, making it difficult for the model to effectively process and interpret the content within an event. Conversely, when the video is segmented into a large number of smaller events, each individual event becomes more understandable, but the challenge shifts to synthesizing information across a substantial number of events. Consequently, it is crucial to determine an optimal event number that allows the model to strike a balance, facilitating both the comprehension of individual events and the integration of information across the entire sequence for optimal learning performance. The best result is achieved when $N_{h}$ is set to 6, as shown in~\cref{ablation3}. {When $N_h$ = 1, which means directly comprehending compound events, matching the setup in the second row of~\cref{ablation1} (without EAFM but with OPH), still enables OPH for efficient learning.}

\noindent\textbf{Influence of MLLM choice}.  As shown in~\cref{mllm_choice}. Using alternative MLLMs may lead to subtle performance variations, we argue that the parameter size is not an absolute metric for evaluating an MLLM's capabilities. Using internVL2-1B instead of LLaVA-Phi-3-3.8B improve EVIS by 0.5\% and 0.2\% $\mathcal{J}$\&$\mathcal{F}$ on MeViS and Ref-Youtube-VOS, respectively. However, under identical InternVL2-1B settings, EVIS outperforms VideoLISA by 5.9\% $\mathcal{J}$\&$\mathcal{F}$ on MeViS. VideoLISA employs an identical configuration to the baseline model in Table~\ref{ablation1} Line 1, utilizing n-frame sampling and InternVL2-1B. This alignment precisely highlights the substantial methodological innovations introduced by EVIS.

\begin{figure}[t]
\centering
\includegraphics[width=0.4\textwidth]{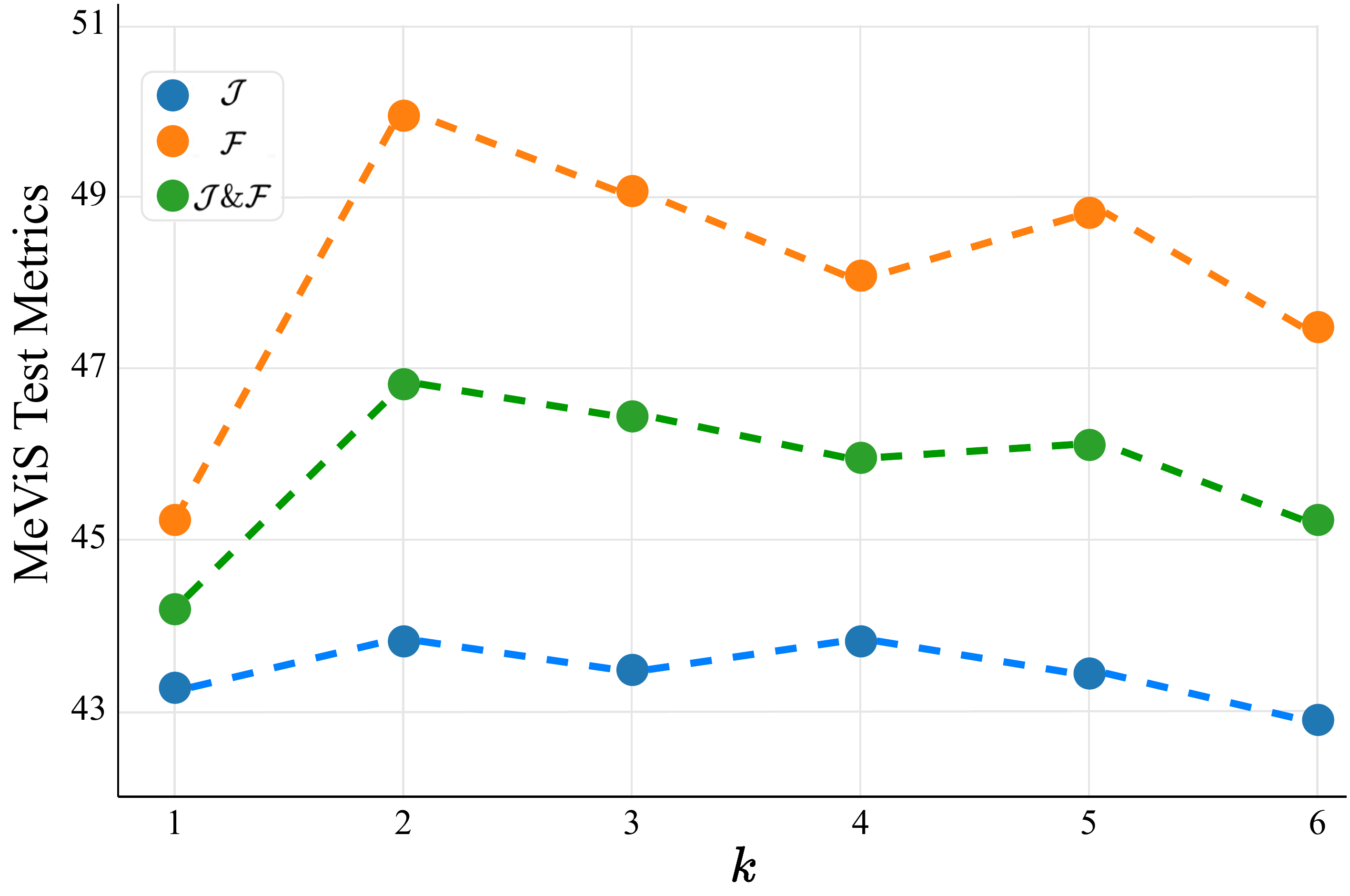}
\caption{Ablation on the effect of top-$k$ event query selection on MeViS~\cite{mevis} dataset. Metrics are region similarity $\mathcal{J}$, contour accuracy $\mathcal{F}$ and their combined average score $\mathcal{J}$\&$\mathcal{F}$, respectively.}
\label{figure6}
\end{figure}

\noindent\textbf{Effect of Top-$k$ Event Query Selection}. In the Frame Merging Block of EAFM, as the number of object queries assigned to an event query increases, the complexity of the information encapsulated within the event query also grows. Conversely, when an event query is associated with fewer object queries, the informational overlap between distinct events becomes minimal, hindering information integration and propagation across consecutive events. As shown in~\cref{figure6}, we conduct ablations on the effect of top-$k$ event query selection. The performance deteriorates when the value of $k$ is either excessively small or overly large. The best result is achieved when the number of $k$ is set to 2.

\begin{table}[t]
    \footnotesize
        \centering
        \caption{Ablation study on MLLM impact.} 
        \setlength\tabcolsep{3pt}
        \fontsize{6}{8}\selectfont
        \renewcommand\arraystretch{1.0}
        \begin{tabular}{l|l|ccc}
            \specialrule{.1em}{.05em}{.05em}
            {Method} & {Multi-Modal LLM} & {MeViS} & {Ref-Ytb-VOS} &{Ref-DAVIS17}  \\
            \hline
            VideoLISA (n-frame) &LLaVA-Phi-3-3.8B &43.2 &63.3 &68.6 \\
            VideoLISA (SDS) &LLaVA-Phi-3-3.8B &44.4 & 63.7 & 68.8 \\
            \textbf{EVIS} &LLaVA-Phi-3-3.8B &46.3 &64.2 &\textbf{69.3} \\
            VideoLISA (n-frame) & InternVL2-1B &39.4 &61.9 &65.7 \\
            VideoLISA (SDS) &InternVL2-1B &40.9 &62.4 &66.4 \\
            \textbf{EVIS} &InternVL2-1B &\textbf{46.8} &\textbf{64.4} &68.8 \\
            \specialrule{.1em}{.05em}{.05em}
        \end{tabular} 
    \label{mllm_choice}
\end{table}

\begin{table}[t]
    \centering
    \setlength\tabcolsep{20pt}
    \caption{Different number of stacking depth $L$ in EAFM on MeViS~\cite{mevis}.}
    \label{ablation_depth}

    \begin{tabular}{c|ccc}
    \specialrule{.1em}{.05em}{.05em}
    \multirow{1}{*}{$L$} & \multirow{1}{*}{$\mathcal{J}$\&$\mathcal{F}$} &\multirow{1}{*}{$\mathcal{J}$}  &\multirow{1}{*}{$\mathcal{F}$} \\
    \hline
         1 &43.1 &39.8 &46.4  \\
         2 &46.4 &43.1 &49.7  \\
         3 &\textbf{46.8} &\textbf{43.7} &\textbf{49.9}  \\
         4 &46.8 &43.8 &49.8  \\
    \specialrule{.1em}{.05em}{.05em}
    \end{tabular}
\end{table}

\begin{table}[t!]
    \centering
    \setlength\tabcolsep{15pt}
    \caption{Different ViT backbone of SAM~\cite{segment-anything} on MeViS~\cite{mevis}.}
    \label{sam_backbone}
  
    \begin{tabular}{c|ccc}
    \specialrule{.1em}{.05em}{.05em}
    \multirow{1}{*}{Visual Encoder} & \multirow{1}{*}{$\mathcal{J}$\&$\mathcal{F}$} &\multirow{1}{*}{$\mathcal{J}$}  &\multirow{1}{*}{$\mathcal{F}$} \\
    \hline
         SAM-ViT-B &42.4 &40.3 &44.5  \\
         SAM-ViT-L &45.3 &43.0 &45.6  \\
         SAM-ViT-H &\textbf{46.8} &\textbf{43.7} &\textbf{49.9}  \\
    \specialrule{.1em}{.05em}{.05em}
    \end{tabular}
\end{table}

\begin{table}[t!]
    \centering
    \caption{Different number of object queries in Mask2Former~\cite{mask2former}.}
    \setlength\tabcolsep{10pt}
    \label{mask2former}
    \begin{tabular}{c|ccc}
    \specialrule{.1em}{.05em}{.05em}
    \multirow{1}{*}{Number of Object Queries} & \multirow{1}{*}{$\mathcal{J}$\&$\mathcal{F}$} &\multirow{1}{*}{$\mathcal{J}$}  &\multirow{1}{*}{$\mathcal{F}$} \\
    \hline
         100 &43.9 &41.0 &46.8  \\
         50 &45.7 &42.6 &48.8  \\
         20 &\textbf{46.8} &\textbf{43.7} &\textbf{49.9}  \\
         10 &46.4 &43.2 &49.6  \\
    \specialrule{.1em}{.05em}{.05em}
    \end{tabular}
\end{table}

\noindent\textbf{Impact of Stacking Depth $L$ in EAFM}. \cref{ablation_depth} shows the performance of EVIS with varying stacking depth $L$ on MeViS~\cite{mevis} dataset. The results indicate that increasing the number of stacked layers leads to improved performance. To balance segmentation capability and computational efficiency, we select $L=3$ as the optimal configuration.

\noindent\textbf{Different Backbones in SAM}. We modify SAM's ViT architecture from H to L and B to evaluate the effectiveness of EVIS, demonstrating its stable progression across different backbone configurations. As shown in~\cref{sam_backbone}, when the backbone of SAM changes from ViT-H to ViT-L and ViT-B, the performance decreases by 1.5\% $\mathcal{J}$\&$\mathcal{F}$ and 4.4\% $\mathcal{J}$\&$\mathcal{F}$, respectively, indicating the effectiveness of our method in fully leveraging the feature information provided by SAM, even with varying backbone complexities.

\noindent\textbf{Ablation studies for Mask2Former.} The number of object queries has a direct impact on the model's segmentation performance, as shown in~\cref{mask2former}, the best result is achieved when number of object query is set to 20. We also conduct ablation studies on whether to freeze Mask2Former. The experimental results demonstrate that when Mask2Former is trained in an unfrozen manner, the performance on MeViS decreases significantly from 46.8\% to 42.3\%. We hypothesize this degradation stems from the disruption of Mask2Former's pre-trained query context representations.

\begin{table}[t!]
    \centering
    \setlength\tabcolsep{5pt}
    \caption{Ablation study on the different training datasets.}
    \label{trainingdata}
  
    \begin{tabular}{cccc|ccc}
    \specialrule{.1em}{.05em}{.05em}
    \multicolumn{4}{c|}{Training Data} & \multicolumn{3}{c}{MeViS} \\
    SemSeg & RIOS & RVOS & ReasonSeg & $\mathcal{J}$ & $\mathcal{F}$ & $\mathcal{J}$\&$\mathcal{F}$ \\
    \hline
    \ding{51} & \textcolor{gray}{\ding{55}}& \textcolor{gray}{\ding{55}}& \textcolor{gray}{\ding{55}}&36.5 &35.3 &37.7  \\
    \ding{51} & \ding{51} & \textcolor{gray}{\ding{55}}& \textcolor{gray}{\ding{55}}&38.2 &36.3 &40.1  \\
    \ding{51} & \ding{51} & \ding{51} & \textcolor{gray}{\ding{55}}&44.7 &42.0 &47.4  \\
    \ding{51} & \ding{51} & \ding{51} & \ding{51} &\textbf{46.8} &\textbf{43.7} &\textbf{49.9}  \\
    \specialrule{.1em}{.05em}{.05em}
    \end{tabular}
\end{table}

\noindent\textbf{Training Datasets}. We conduct ablation experiments on the training data to assess the influence of various datasets on the proposed approach EVIS, as presented in~\cref{trainingdata}. The addition of RIOS, RVOS (shown in~\cref{trainingdata}), and ReasonSeg to the training set leads to respective improvements of 1.7\%, 6.5\%, and 2.1\% $\mathcal{J}$\&$\mathcal{F}$, respectively.

\begin{table*}[t]
	\centering
	\footnotesize
        \caption{Quantitative evaluation results on MeViS~\cite{mevis}, Ref-Youtube-VOS~\cite{refer-ytb-vos}, and Ref-DAVIS17~\cite{ref-dacvis17}. $\dagger$ indicates our model re-implemented by removing the LLM. Bold indicates the best scores.}
        \setlength\tabcolsep{7.5pt}%
		\renewcommand\arraystretch{1.0}
		\begin{tabular}{r|r|l|ccc|ccc|ccc}
			\specialrule{.1em}{.05em}{.05em}
			\multirow{2}{*}{Method} & \multirow{2}{*}{Reference} & \multirow{2}{*}{Backbone} &\multicolumn{3}{c|}{MeViS} & \multicolumn{3}{c|}{Ref-Youtube-VOS} &\multicolumn{3}{c}{Ref-DAVIS17}  \\
			\cline{4-12}
			
    		& & &$\mathcal{J}$\&$\mathcal{F}$ & $\mathcal{J}$ & $\mathcal{F}$ & $\mathcal{J}$\&$\mathcal{F}$ & $\mathcal{J}$ & $\mathcal{F}$ & $\mathcal{J}$\&$\mathcal{F}$ & $\mathcal{J}$ & $\mathcal{F}$ \\ 
			\hline
            \rowcolor{mygray}
            \multicolumn{12}{l}{\emph{Traditional methods}} \\
            \hline
                URVOS~\cite{refer-ytb-vos} &ECCV'20 &ResNet-50 &27.8 &25.7 &29.9 &47.2 &45.2 &49.2 &51.6 &47.3 &55.9 \\
    		LBDT~\cite{lbdt} &CVPR'22 &ResNet-50 &29.3 &27.8 &30.8 &49.4 &48.2 &50.6 &54.3 &- &- \\
                MLSA~\cite{mlsa} &CVPR'22 &ResNet-50 &- &- &-&49.7 & 48.4 & 50.9 & 57.9 & 53.8 & 62.0 \\
                MTTR~\cite{mttr}  &CVPR'22 &Video-Swin-T &30.0 &28.8 &31.2 & 55.3 & 54.0 & 56.6 & - & - & - \\
    		ReferFormer~\cite{referformer}  &CVPR'22 & Video-Swin-B&31.0 &29.8 &32.2 & 62.9 & 61.3 & 64.6 & 61.1 &58.1 &64.1 \\
                HTML~\cite{html} &ICCV'23 &Video-Swin-B &- &- &- &63.4 &61.5 &65.2 &62.1 &59.2 &65.1  \\
                R$^2$-VOS~\cite{r2vos} &ICCV'23 &Video-Swin-T &- &- &- &61.3 &59.6 &63.1 &- &- &- \\ 
    		SgMg~\cite{sgmg} &ICCV'23 &Video-Swin-T &- &- &- & 62.0 & 60.4 & 63.5 & 61.9 & 59.0 & 64.8 \\
    		OnlineRefer~\cite{onlinerefer} &ICCV'23 &Swin-B&- &- &- & 62.9 & 61.0 & 64.7 & 62.4 & 59.1 & 65.6 \\
                TempCD~\cite{tempcd} &ICCV'23 &Video-Swin-T &- &- &- &62.3 &60.5  &64.0  &62.2  &59.3  &65.0  \\
                SOC~\cite{soc} &NeurIPS'23 &Video-Swin-T &- &- &- &62.4  &61.1  &63.7  &63.5  &60.2  &66.7  \\
                VLT+TC~\cite{vlt+tc} &TPAMI'23 &Video-Swin-B &35.5 &33.6 &37.3 &63.8  &61.9  &65.6  &61.6  &58.9  &64.3  \\
                LMPM~\cite{mevis} &ICCV'23  & Swin-T&37.2 & 34.2& 40.2& - & - & - & - & - & - \\
                LoSh~\cite{losh} &CVPR'24 & Swin-T &- &- &- &63.7 &62.0 &65.4 &62.9 &60.1 &65.7 \\
                DsHmp~\cite{dshmp} &CVPR'24 & Swin-T/Video-Swin-T &46.4 & 43.0& 49.8& 63.6 & 61.8 & 65.4 & 64.0 & 60.8 & 67.2 \\ 
                \textbf{EVIS$\dagger$} (ours) & -\ \ \ \ \ \ \ \ \ \  & Swin-T &\textbf{46.7} &\textbf{43.5} &\textbf{49.9} &\textbf{64.1} &\textbf{62.2} &\textbf{66.0} &\textbf{65.6} &\textbf{62.0} &\textbf{69.2} \\
            \hline
            \rowcolor{mygray}
            \multicolumn{12}{l}{\emph{LLM-based methods}} \\
            \hline
			LISA-7B~\cite{lisa} &CVPR'23 &LLaVA-7B &37.2 &35.1 &39.4 & 50.2 & 49.7 & 50.6 & 58.4 & 54.9 & 61.9 \\
			TrackGPT-7B~\cite{trackgpt} &arXiv'23 &LLaVA-7B &40.1 &37.6 &42.6 &56.4 &55.3 &57.4 &63.2  &59.4  &67.0 \\
                VISA-7B~\cite{visa} & ECCV'24 &Chat-UniVi-7B &43.5 &40.7 &46.3 & 61.5 & 59.8 & 63.2 & \textbf{69.4} & \textbf{66.3} & 72.5 \\
                VideoLISA-3.8B~\cite{videolisa} & NeurIPS'24 &LLaVA-Phi-3-3.8B &44.4 &41.3 &47.6 & 63.7 & 61.7 &65.7 & 68.8 & 64.9 & \textbf{72.7} \\
                \textbf{EVIS-1B} (ours) & -\ \ \ \ \ \ \ \ \ \ &InternVL2-1B &\textbf{46.8} &\textbf{43.7} &\textbf{49.9} &\textbf{64.4} &\textbf{62.6} &\textbf{66.2} &68.8 &65.8 &71.8 \\
            \specialrule{.1em}{.05em}{.05em}
		\end{tabular} 
	\label{table:sota_refer}
\end{table*}

\begin{table}[t!]
\footnotesize
\centering
\caption{Results on A2D-Sentences and JHMDB-Sentences.} \label{tab:A2D_JHMDB}
\setlength{\tabcolsep}{4.5pt}
\begin{tabular}{r| c c c | c c c }
\specialrule{.1em}{.05em}{.05em}
& \multicolumn{3}{c |}{A2D-Sentences~\cite{a2d-sentence}} & \multicolumn{3}{c}{JHMDB-Sentences~\cite{jhmdb}} \\
 Method& mAP &oIoU & mIoU & mAP & oIoU & mIoU \\
\hline
ReferFormer~\cite{referformer} &  55.0 & 78.6 & 70.3 & 43.7 & 73.0 & 71.8\\
OnlineRefer~\cite{onlinerefer} & -&79.6& 70.5&- &73.5 &71.9\\
HTML~\cite{html}& 56.7 & 79.5 &71.2 &44.2&-&-\\ 
SOC~\cite{soc}&57.3  &80.7 & 72.5 & 44.6 &73.6 &72.3\\
SgMg~\cite{sgmg} &  {58.5} & {79.9} & {72.0} & {45.0} & {73.7} & {72.5}  \\ 
DsHmp~\cite{dshmp}&59.8&81.1&72.9&45.8&73.9&73.0\\
LoSh~\cite{losh} &59.9 &81.2 &73.1 &45.7 &74.5 &73.4 \\
\hline
\textbf{EVIS} (\textbf{ours}) &\textbf{60.5}&\textbf{81.3}&\textbf{73.5}&\textbf{46.2}&\textbf{75.5}&\textbf{74.2}\\
\specialrule{.1em}{.05em}{.05em}
\end{tabular}
\end{table}
\begin{table}[t!]
\footnotesize
\centering
\caption{Reasoning segmentation results on ReVOS~\cite{visa}.} \label{tab:ReasonVOS}
\setlength{\tabcolsep}{6.3pt}
\begin{tabular}{r|l|ccc}
\specialrule{.1em}{.05em}{.05em}
Method & Backbone & $\mathcal{J}$\&$\mathcal{F}$ & $\mathcal{J}$ & $\mathcal{F}$ \\
\hline
MTTR~\cite{mttr}& Video-Swin-T &21.0 &20.4 &21.5 \\
LMPM~\cite{mevis}& Swin-T &18.8 &13.3 &24.3 \\
ReferFormer~\cite{referformer}& Video-Swin-B &23.4 &21.3 &25.6 \\
LISA~\cite{lisa}& LLaVA-7B &36.1 &33.8 &38.4 \\
TrackGPT~\cite{trackgpt}& LLaVA-7B &39.0 &36.8 &41.2 \\
VISA~\cite{visa} & Chat-UniVi-7B &39.2 &36.7 &41.7 \\
VideoLISA~\cite{videolisa} & LLaVA-Phi-3-3.8B &39.5 &37.3 &41.7 \\
\hline
\textbf{EVIS} (\textbf{ours}) &InternVL2-1B &\textbf{40.3} &\textbf{38.0} &\textbf{42.6}\\
\specialrule{.1em}{.05em}{.05em}
\end{tabular}
\end{table}
\begin{table}[t!]
    \centering
    \footnotesize
    \setlength{\tabcolsep}{2.5pt}
    \caption{Reasoning segmentation results on ReasonSeg~\cite{lisa} benchmark.} \label{table:resonseg}
    \begin{tabular}{ l | c c | c c | c c | c c }
    \specialrule{.1em}{.05em}{.05em}            
        & \multicolumn{2}{c|}{val} & \multicolumn{6}{c}{test} \\ 
        \cline{2-9}
        & \multicolumn{2}{c|}{overall} & \multicolumn{2}{c|}{short query} & \multicolumn{2}{c|}{long query} & \multicolumn{2}{c}{overall} \\
        Method & gIoU & cIoU & gIoU & cIoU & gIoU & cIoU & gIoU & cIoU \\ 
        \hline
        OVSeg~\cite{ovseg} & 28.5 & 18.6 & 18.0 & 15.5 & 28.7 & 22.5 & 26.1 & 20.8  \\
        RELA~\cite{GRES} & 22.4 & 19.9 & 17.6 & 15.0 & 22.6 & 23.8 & 21.3 & 22.0 \\    %
        X-Decoder~\cite{x-decoder} & 22.6 & 17.9 & 20.4 & 11.6 & 22.2 & 17.5 & 21.7 & 16.3 \\
        SEEM~\cite{SEEM} & 25.5 & 21.2 & 20.1 & 11.5 & 25.6 & 20.8 & 24.3 & 18.7 \\
        Grounded-SAM~\cite{grounded-sam} & 26.0 & 14.5 & 17.8 & 10.8 & 22.4 & 18.6 & 21.3 & 16.4 \\
        LISA-7B~\cite{lisa} & 44.4 & 46.0 & 37.6 & 34.4 & 36.6 & 34.7 & 36.8 & 34.1 \\
        VISA-7B~\cite{visa} &52.7 &57.8 &-  &-  &-  &-  &-  &-  \\
        VideoLISA-3.8B~\cite{videolisa} & \textbf{61.4} & \textbf{67.1} & \textbf{43.8} & \textbf{42.7} & \textbf{56.9} & \textbf{57.7} & \textbf{53.8} & \textbf{54.4} \\
        \hline
        \textbf{EVIS-1B} (\textbf{ours}) & 55.4  &61.3  &42.5  &43.0  &44.3  &43.7  &43.8  &44.6 \\
    \specialrule{.1em}{.05em}{.05em}     
    \end{tabular}
\end{table}
\begin{table}[ht]
\footnotesize
\centering
\caption{Quantitative evaluation results (cIoU) on RefCOCO, RefCOCO+~\cite{refcoco+} and RefCOCOg~\cite{refcocog}. Bold indicates the best scores.} 
\label{table:rios}
\setlength{\tabcolsep}{1.9pt}
\begin{tabular}{l| c c c | c c c | c c }
\specialrule{.1em}{.05em}{.05em}
& \multicolumn{3}{c|}{RefCOCO~\cite{refcoco+}} & \multicolumn{3}{c|}{RefCOCO+~\cite{refcoco+}} & \multicolumn{2}{c}{RefCOCOg~\cite{refcocog}}\\
 Method &val &testA &testB &val &testA &testB &val(U) &test(U)\\
\hline
MCN~\cite{mcn} & 62.4 & 64.2 & 59.7 & 50.6 & 55.0 & 44.7 & 49.2 & 49.4\\
VLT~\cite{vlt} & 67.5 & 70.5 & 65.2 & 56.3 & 61.0 & 50.1 & 55.0 & 57.7\\
CRIS~\cite{cris} & 70.5 & 73.2 & 66.1 & 62.3 & 68.1 & 53.7 & 59.9 & 60.4\\
LAVT~\cite{LAVT} & 72.7 & 75.8 & 68.8 & 62.1 & 68.4 & 55.1 & 61.2 & 62.1\\
ReLA~\cite{GRES} & 73.8 & 76.5 & 70.2 & \textbf{66.0} & \textbf{71.0} & \textbf{57.7} & 65.0 & 66.0\\
X-Decoder~\cite{x-decoder} & - & - & - & - & - & - & 64.6 & -\\
SEEM~\cite{SEEM} & - & - & - & - & - & - & 65.7 & -\\
VISA-7B~\cite{visa} & 72.4  &75.5  &68.1  &59.8 &64.8  &53.1  &65.5  &66.4 \\
LISA-7B~\cite{lisa} & 74.1 & 76.5 & \textbf{71.1} & 62.4 & 67.4 & 56.5 & 66.4 & 68.5\\
VideoLISA-3.8B~\cite{videolisa} & 73.8 & 76.6 & 68.8 & 63.4 & 68.8 & 56.2 & 68.3 & 68.8\\
\hline
\textbf{EVIS-1B} (\textbf{ours}) &\textbf{74.2}  &\textbf{76.8}  &68.9  &63.9  &70.1  &56.2  &\textbf{68.5}  &\textbf{69.7} \\
\specialrule{.1em}{.05em}{.05em}
\end{tabular}
\end{table}

\subsection{Comparison with State-of-the-Art Methods}
With event-level analysis, EVIS achieves remarkable performance across both traditional and LLM-based methods.

\noindent\textbf{MeViS}~\cite{mevis}.
In~\cref{table:sota_refer}, we evaluate the proposed EVIS on the newly released MeViS dataset for the referring video segmentation task. Following the experimental setup in~\cite{videolisa}, our approach outperforms existing state-of-the-art methods, achieving a significant improvement of 2.4\% in $\mathcal{J}$\&$\mathcal{F}$ over the previous LLM-based best-performing method, VideoLISA~\cite{videolisa}. Notably, our model is based on an MLLM with only 1B parameters, significantly fewer than the VideoLISA-3.8B and VISA-13B models, further validating the effectiveness of event-by-event learning. Furthermore, EVIS$\dagger$ also achieves competitive performance on MeViS, demonstrating the EAFM module is compatible with both LLM-based and traditional methods without reasoning capability.

\begin{figure*}[t]
\centering
\includegraphics[width=0.98\textwidth]{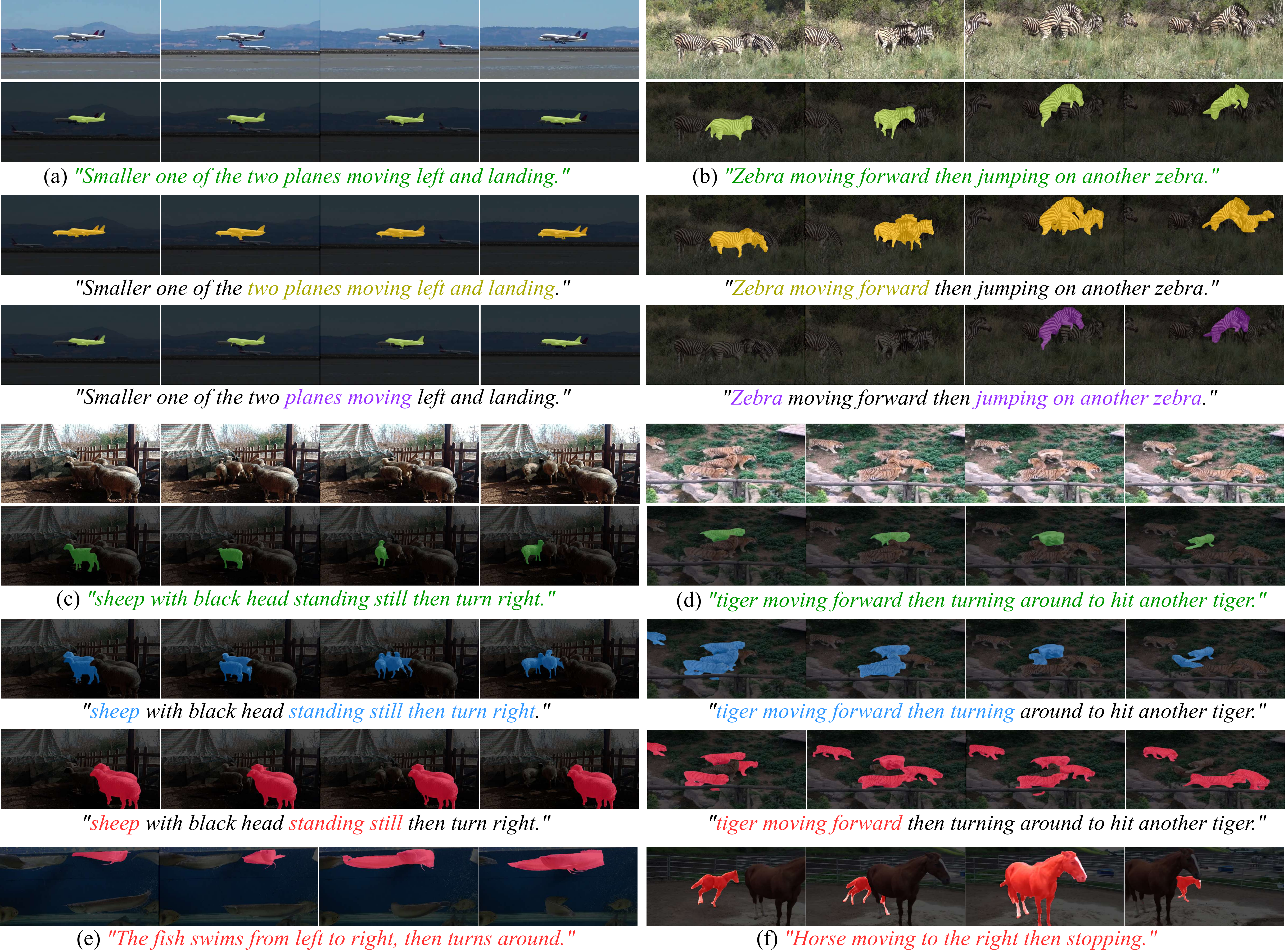}
\caption{Example success and failure cases of EVIS. The black font denotes text that is not visible to the model.}
\label{figure8}
\end{figure*}

\begin{figure*}[!t]
\centering
\includegraphics[width=0.98\textwidth]{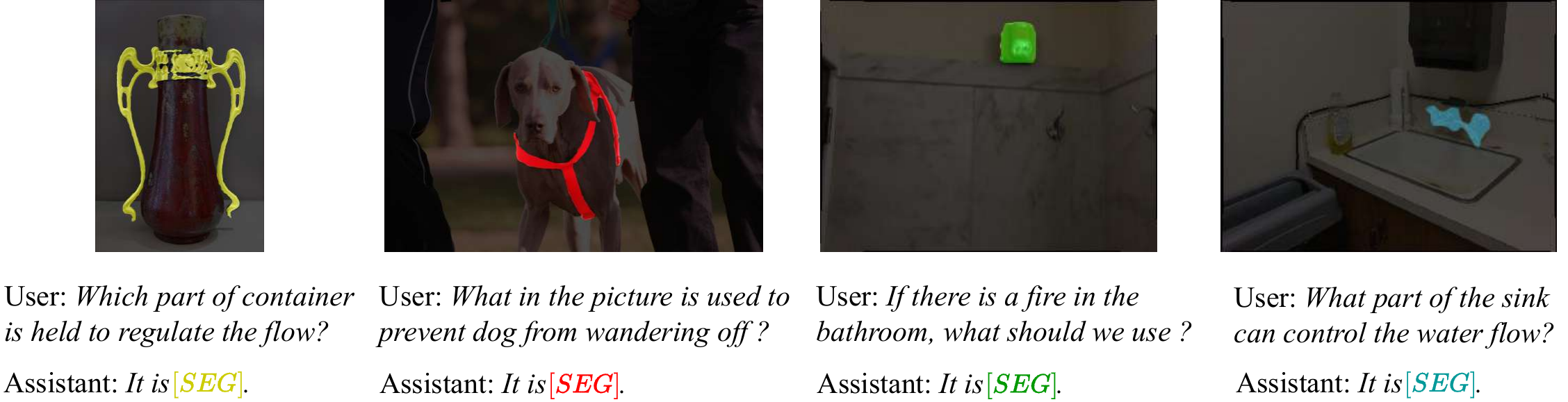}
\caption{Qualitative results of EVIS on ReasonSeg~\cite{lisa} Dataset.}
\label{figure11}
\end{figure*}

\noindent\textbf{Ref-YouTube-VOS}~\cite{refer-ytb-vos} \textbf{and} \textbf{Ref-DAVIS17}~\cite{ref-dacvis17}.
In~\cref{table:sota_refer}, we present results on Ref-YouTube-VOS and Ref-DAVIS17 datasets. Our method outperforms most approaches across all evaluation metrics. On Ref-YouTube-VOS, EVIS with the InternVL-1B achieves a $\mathcal{J}$\&$\mathcal{F}$ score of 64.4\%, surpassing VideoLISA-3.8B~\cite{videolisa} by 0.7\%. On the Ref-DAVIS17 dataset, EVIS achieves a 68.8\% in $\mathcal{J}$\&$\mathcal{F}$, surpassing the state-of-the-art traditional method~\cite{dshmp} by 4.8\%. It is crucial to emphasize that VISA-7B~\cite{visa} achieves only a marginal 0.6\% $\mathcal{J}$\&$\mathcal{F}$ improvement over the proposed EVIS-1B. However, VISA employs the Chat-UniVi-7B~\cite{chatunivi} as its MLLM, which has a significantly larger number of parameters compared to our method. These results underscore the effectiveness of EVIS in complex scenarios.

\noindent\textbf{A2D-Sentences}~\cite{a2d-sentence} \textbf{and} \textbf{JHMDB-Sentences}~\cite{jhmdb}. We follow~\cite{dshmp} to fine-tune the model on A2D-Sentences. In~\cref{tab:A2D_JHMDB}, EVIS achieves new state-of-the-art results, outperforming the LoSh~\cite{losh}, by 0.6\% and 0.5\% mAP on the A2D-Sentences and JHMDB-Sentences datasets, respectively. The relatively smaller performance improvements on these datasets, compared to MeViS, can be attributed to the presence of simpler, image-level descriptions in the sentences, which lack the complexity of event variations.

\noindent\textbf{RefCOCO}, \textbf{RefCOCO+}~\cite{refcoco+} \textbf{and} \textbf{RefCOCOg}~\cite{refcocog}. We evaluate the proposed EVIS on three benchmarks for the referring image segmentation task. As shown in~\cref{table:rios}, while our original focus is on addressing the referring video segmentation task, our approach demonstrates competitive performance on image tasks, highlighting the effectiveness of our method.

\noindent\textbf{ReVOS}~\cite{visa}. Reasoning Video Object Segmentation (ReasonVOS) is a new and challenging task that differs from RVOS in that it requires the ability to reason using world knowledge. In RVOS, model might be tasked with identifying \textit{“a running cat”}, whereas in ReasonVOS, it needs to identify \textit{“the smartest cat”}. We evaluate EVIS's reasoning ability on the ReVOS~\cite{visa} dataset, as shown in~\cref{tab:ReasonVOS}. Notably, EVIS is not trained on any Video Reasoning datasets, yet it still outperforms VideoLISA~\cite{videolisa} by 0.8\% $\mathcal{J}$\&$\mathcal{F}$.

\noindent\textbf{ReasonSeg}~\cite{lisa}. Reasoning capabilities are a hallmark of LLMs, enabling them to capture complex relationships and contextual dependencies. To validate the image reasoning effectiveness of our approach, we conduct a comprehensive comparison with state-of-the-art methods on the ReasonSeg~\cite{lisa} benchmark. As shown in ~\cref{table:resonseg}, our method achieves competitive performance across multiple metrics, demonstrating its superior ability to handle complex segmentation scenarios. These results demonstrate the robustness and generalizability of EVIS, confirming its advantages over traditional segmentation approaches. Qualitative results in~\cref{figure11}.

\begin{figure*}[t]

\centering
\includegraphics[width=0.94\textwidth]{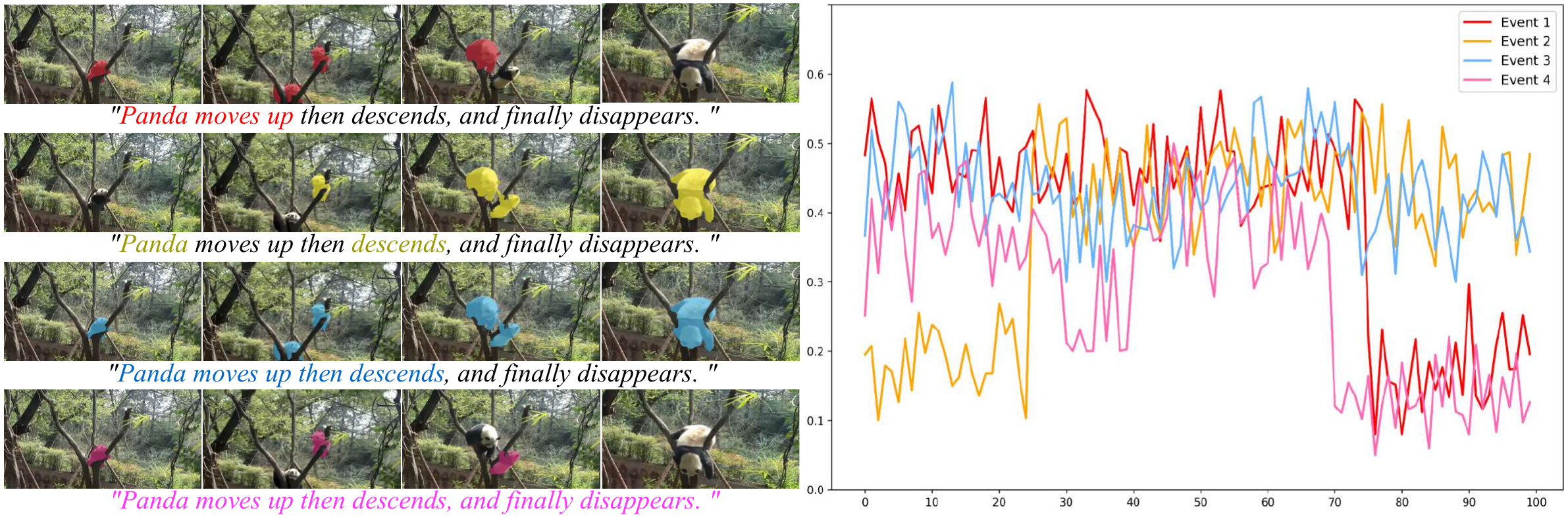}
\vspace{-2mm}
\caption{Visualization for event decomposition. Scores are calculated by cosine similarity between global object and event queries in EAFM.}
\vspace{-3mm}
\label{figure:sim}
\end{figure*}

\begin{figure}[t]

\centering
\includegraphics[width=0.48\textwidth]{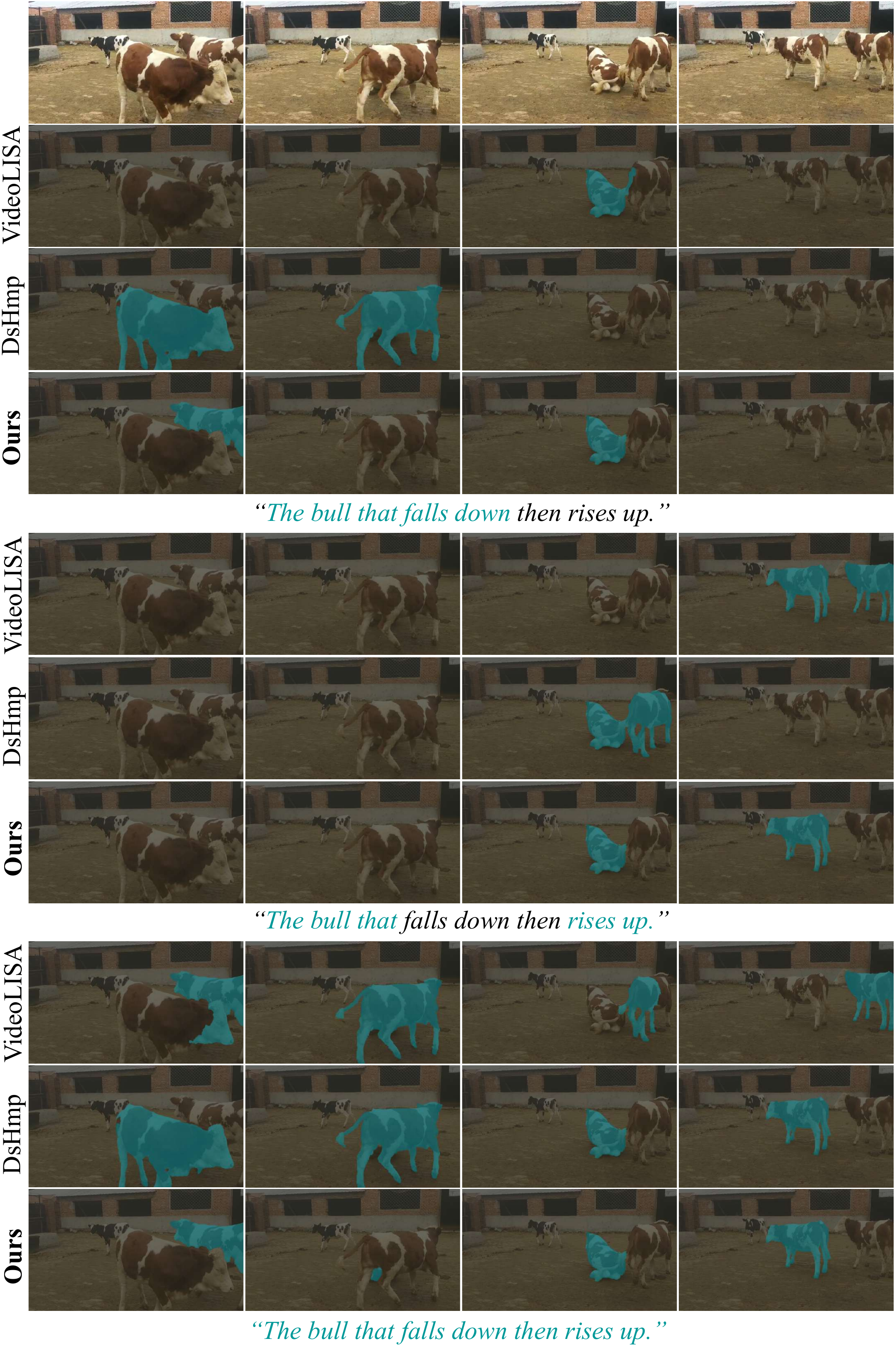}
\caption{Qualitative comparisons with VideoLISA~\cite{videolisa} and DsHmp~\cite{dshmp}.}
\vspace{-3mm}
\label{figure10}
\end{figure}

\begin{figure}[t]
\centering
\includegraphics[width=0.48\textwidth]{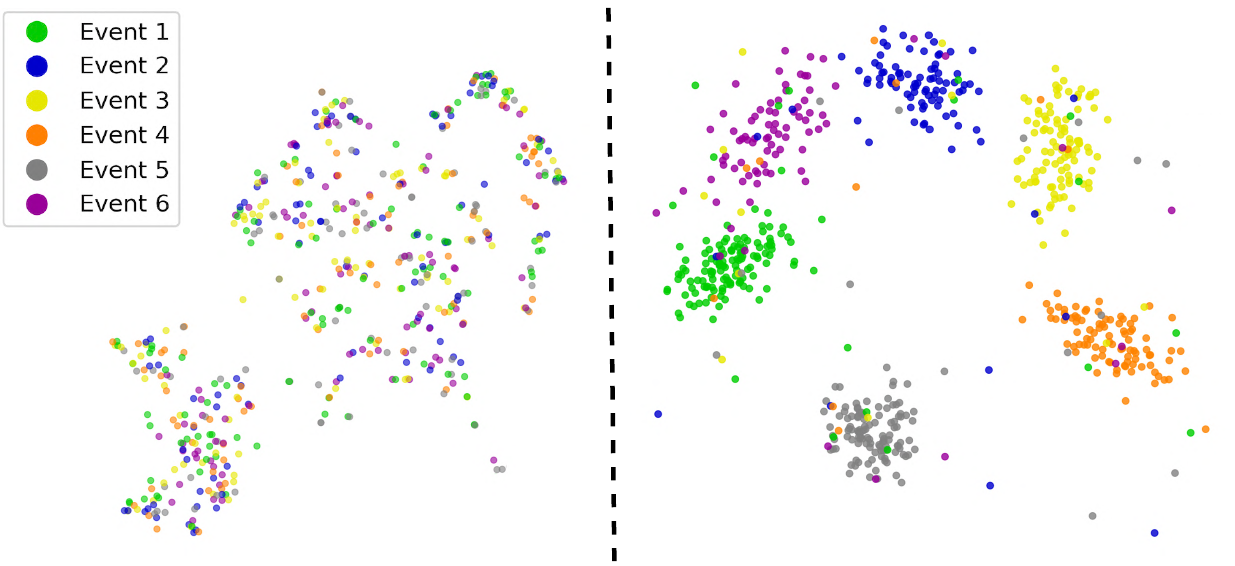}
\caption{Visulization of global queries learned w/o training (left) and w/ training (right). Features are colored according to the similarity between global object and event queries. Best viewed in color.}
\vspace{-3mm}
\label{figure9}
\end{figure}

\noindent\textbf{Training and inference costs comparisons}. We conduct experiments using a single NVIDIA A6000 GPU on MeViS dataset. As shown in~\cref{table:infer_costs}, Compared to VideoLISA, EVIS has 58\% fewer parameters and runs at 3.6× the inference speed. Furthermore, EVIS significantly outperforms DsHmp in terms of trainable parameters, computational complexity, and inference speed. The trainable parameters of EAFM and OPH are approximately 14.45M, making them extremely lightweight compared to most existing models. {We also conduct experiments on EAFM and OPH, as shown in~\cref{table:trade-off}.}

\begin{table}[t]
    \centering
    \caption{Training and inference costs comparisons.} 
        \setlength\tabcolsep{1.4pt}
        \fontsize{7.5}{9.0}\selectfont
        \renewcommand\arraystretch{1.0}
        \begin{tabular}{l|l|lll}
            \specialrule{.1em}{.05em}{.05em}
            {Method} & {Backbone} & {Training Params (M)} & {FLOPs (G)}& {Inference (FPS)} \\
            \hline
            VideoLISA & LLaVA-Phi-3-3.8B &$\approx$ 380 &$\approx$ 8.25 & $\approx$ 9.8 \\
            {EVIS} & {LLaVA-Phi-3-3.8B} & {$\approx$} {395}  & {$\approx$} {8.47}  & {$\approx$} {9.2} \\
            \textbf{EVIS} &InternVL2-1B & $\approx$ \textbf{160}  &$\approx$ \textbf{3.15}  & $\approx$ \textbf{35.4} \\
            DsHmp &Swin-T &$\approx$ 28 &$\approx$ 0.21 & $\approx$ 58.8 \\
            \textbf{EVIS}$\dagger$ &Swin-T &$\approx$ \textbf{20} &$\approx$ \textbf{0.19} & $\approx$  \textbf{65.5 }\\
            {EVIS} {$\dagger$} & {Swin-B} & {$\approx$} {63} & {$\approx$} {0.70}& {$\approx$}  {31.5} \\
            {EVIS} {$\dagger$} & {Swin-L} & {$\approx$}  {142} & {$\approx$} {1.58} & {$\approx$}  {12.4} \\
            \specialrule{.1em}{.05em}{.05em}
        \end{tabular} 
    \label{table:infer_costs}
\end{table}

\begin{table}[!ht]
    \centering
    \setlength\tabcolsep{5pt}
    \renewcommand\arraystretch{1.1}

    \caption{Ablation study on different configurations of EVIS components on the MeViS dataset. Params represent the total number of parameters in the EAFM and OPH modules.}
    \begin{tabular}{cc|cc|cccc}
    \specialrule{.1em}{.05em}{.05em}
    \multicolumn{2}{c|}{\textbf{EAFM}} & \multicolumn{2}{c|}{\textbf{OPH}} & \multirow{2}{*}{Params (M)} & \multirow{2}{*}{FLOPs (G)} & \multirow{2}{*}{FPS} & \multirow{2}{*}{$\mathcal{J}$\&$\mathcal{F}$} \\
    \cline{1-4}
    $L$ & $N_h$ & $T_u$ & $T_o$ & & & & \\    \hline
    \textbf{3} & \textbf{6} & \textbf{4} & \textbf{8} & $\approx$ \textbf{14.45} & $\approx$ \textbf{0.14} & \textbf{35.4} &\textbf{46.8}  \\
    4 & 6 & 4 & 8 & $\approx$ 19.17 & $\approx$ 0.20 & 34.5 &46.5 \\
    5 & 6 & 4 & 8 & $\approx$ 23.89 & $\approx$ 0.23 & 33.9 &46.1 \\
    \hline
    3 & 6 & 2 & 4 & $\approx$ 8.33 & $\approx$ 0.09 & 37.1 &41.5 \\
    \textbf{3} & \textbf{6} & \textbf{4} & \textbf{8} & $\approx$ \textbf{14.45} & $\approx$ \textbf{0.14} & \textbf{35.4} &\textbf{46.8}  \\
    3 & 6 & 6 & 12 & $\approx$ 20.70 & $\approx$ 0.19 & 31.6 &46.7 \\
    3 & 6 & 8 & 16 & $\approx$ 27.27 & $\approx$ 0.27 & 25.8 &45.6 \\
    \specialrule{.1em}{.05em}{.05em}
    \end{tabular}
    \label{table:trade-off}
\end{table}

\noindent\textbf{Qualitative Results}. \cref{figure8} displays some cases of the proposed EVIS. Example (a), (b), (c) and (d) show successful cases where EVIS accurately interprets partial phrase and full expressions, demonstrating its event-level video comprehension. Examples (e) and (f) are failure cases. In case (e), the target fish, temporarily occludes the head of another fish, causing the model to perceive them as a single object. Similarly, in case (f), the target horse is obscured by another horse, resulting in temporal occlusion that challenges the model's reasoning and self-correction capabilities.
To further illustrate the model's capability, we also compare EVIS with SOTA methods, including VideoLISA and DsHmp, in~\cref{figure10}. {We observe that prior methods struggle dramatically with queries involving multi-stage temporal dynamics. In contrast, our method correctly identifies and tracks the target throughout the partial or entire event sequence.}

{
\noindent\textbf{Event Decomposition Visualization}. We visualize cosine similarity between global object and event queries in EAFM, showing similarity dynamics across frames with frame numbers and similarity scores, which verify that our approach captures event-level semantics, as shown in~\cref{figure:sim}.
}

\noindent\textbf{t-SNE Visualization}. We Visualize global queries learned w/o training and w/ training. As shown in~\cref{figure9}, features are colored according to the similarity between global object and event queries. Initially, the global queries exhibit a lack of clear structure. However, after training, the event-level queries form well-separated clusters. This demonstrates EVIS's capacity to learn hierarchical event semantics, rather than simply partitioning the video into temporal segments.

\section{Limitations}
\label{sec:limitations}

\noindent\textbf{Object Occlusion}. As shown in~\cref{figure8}~(f), excessive mutual occlusion, especially when objects are fully obscured, can lead to errors in model judgment. Current methods lack the ability to reason about absent objects, highlighting the need to incorporate spatial reasoning and temporal context.

\noindent\textbf{Segmentation Paradigm of LLMs}. Existing LLM-based approaches typically follow the LISA~\cite{lisa} framework, which extends the LLM’s vocabulary by introducing a \texttt{[SEG]} token and employs the embedding-as-mask paradigm to facilitate segmentation. However, the features encoded by the \texttt{[SEG]} token capture limited coverage of the language model’s overall output. To fully leverage the reasoning capabilities of LLMs, it's essential to enhance the utilization of hidden features. Furthermore, exploring new paradigms for LLM-based segmentation remains a promising avenue for future research.

\section{Conclusion}
\label{sec:conclusion}
We propose EVIS, an Event-Aware Video Instructed Segmentation Assistant, which leverages object features from simple events to extract event-aware object trajectories, enabling a hierarchical understanding of videos. We further introduce the Event-Aware Frame Merging Module (EAFM), which utilizes text-guided event queries to merge objects from multiple frames, generating distinct simple events. To enhance the model's learning capabilities, we design event-intra attention for fine-grained learning within events, and event-inter attention for long-term learning across events, ensuring effective alignment between video content and its corresponding expression. Moreover, we propose Object-Pixel-Mixing Learning, a strategy that accelerates target tracking over long temporal spans by integrating fine-grained pixel-level features with prior object tokens, thus enhancing the performance of MLLMs. Our approach demonstrates significant advancements in accuracy and efficiency in referring video segmentation.

\bibliographystyle{IEEEtran}
\bibliography{main}  

@String(PAMI = {IEEE Trans. Pattern Anal. Mach. Intell.})

@String(CVPR= {IEEE Conf. Comput. Vis. Pattern Recog.})

@String(ICCV= {Int. Conf. Comput. Vis.})

@String(ECCV= {Eur. Conf. Comput. Vis.})

@String(NeurIPS= {Adv. Neural Inform. Process. Syst.})

@String(BMVC= {Brit. Mach. Vis. Conf.})

@String(TMM  = {IEEE Trans. Multimedia})

@String(ACCV  = {ACCV})

@String(ICLR = {Int. Conf. Learn. Represent.})

@String(AAAI = {AAAI})

@book{koffka2013principles,
  title={Principles of Gestalt psychology},
  author={Koffka, Kurt},
  year={2013},
  publisher={routledge}
}

@article{atkinson1968human,
  title={Human memory: A proposed system and its control processes},
  author={Atkinson, Richard C},
  journal={The psychology of learning and motivation},
  volume={2},
  year={1968}
}

@inproceedings{GRES,
  title={{GRES}: Generalized Referring Expression Segmentation},
  author={Liu, Chang and Ding, Henghui and Jiang, Xudong},
  booktitle=CVPR,
  pages={23592--23601},
  year={2023}
}

@inproceedings{LAVT,
  title={LAVT: Language-Aware Vision Transformer for Referring Image Segmentation},
  author={Yang, Zhao and Wang, Jiaqi and Tang, Yansong and Chen, Kai and Zhao, Hengshuang and Torr, Philip HS},
  booktitle=CVPR,
  pages={18155--18165},
  year={2022}
}

@inproceedings{Chen_lang2seg_2019,
  author = {Y.-W. Chen and Y.-H. Tsai and T. Wang and Y.-Y. Lin and M.-H. Yang},
  booktitle = BMVC,
  title = {Referring Expression Object Segmentation with Caption-Aware Consistency},
  year = {2019}
}

@book{event,
  title={Understanding events: From perception to action},
  author={Shipley, Thomas F and Zacks, Jeffrey M},
  year={2008},
  publisher={Oxford University Press},
}

@inproceedings{mevis,
  author = {Henghui Ding and
                  Chang Liu and
                  Shuting He and
                  Xudong Jiang and
                  Chen Change Loy},
  title = {{MeViS}: {A} Large-scale Benchmark for Video Segmentation with Motion Expressions},
  booktitle = ICCV,
  year = {2023},
  pages = {2694--2703}
}

@inproceedings{videolisa,
      title={One Token to Seg Them All: Language Instructed Reasoning Segmentation in Videos}, 
      author={Zechen Bai and Tong He and Haiyang Mei and Pichao Wang and Ziteng Gao and Joya Chen and Lei Liu and Zheng Zhang and Mike Zheng Shou},
      booktitle = NeurIPS,
      year={2024},
      pages={6833--6859}
}

@inproceedings{visa,
      title={VISA: Reasoning Video Object Segmentation via Large Language Models}, 
      author={Cilin Yan and Haochen Wang and Shilin Yan and Xiaolong Jiang and Yao Hu and Guoliang Kang and Weidi Xie and Efstratios Gavves},
      booktitle = ECCV,
      year={2024},
      pages={98--115}
}

@inproceedings{dshmp,
      title={Decoupling Static and Hierarchical Motion Perception for Referring Video Segmentation}, 
      author={Shuting He and Henghui Ding},
      booktitle = CVPR,
      year={2024},
      pages={13332-13341}
}

@inproceedings{lisa,
      title={LISA: Reasoning Segmentation via Large Language Model}, 
      author={Xin Lai and Zhuotao Tian and Yukang Chen and Yanwei Li and Yuhui Yuan and Shu Liu and Jiaya Jia},
      booktitle = CVPR,
      year={2024},
      pages={9579-9589}
}

@inproceedings{llava,
      title={Visual Instruction Tuning}, 
      author={Liu, Haotian and Li, Chunyuan and Wu, Qingyang and Lee, Yong Jae},
      booktitle = NeurIPS,
      year={2023},
      pages={34892--34916}
}

@inproceedings{blip2,
  author       = {Junnan Li and
                  Dongxu Li and
                  Silvio Savarese and
                  Steven C. H. Hoi},
  title        = {{BLIP-2:} Bootstrapping Language-Image Pre-training with Frozen Image Encoders and Large Language Models},
  booktitle    = {{ICML}},
  year         = {2023},
  pages={19730--19742}
}

@inproceedings{minigpt-4,
  author       = {Deyao Zhu and
                  Jun Chen and
                  Xiaoqian Shen and
                  Xiang Li and
                  Mohamed Elhoseiny},
  title        = {MiniGPT-4: Enhancing Vision-Language Understanding with Advanced Large Language Models},
  booktitle    = ICLR,
  year         = {2024},
  pages={18378--18394}
}

@article{internvl,
  title={InternVL: Scaling up Vision Foundation Models and Aligning for Generic Visual-Linguistic Tasks},
  author={Chen, Zhe and Wu, Jiannan and Wang, Wenhai and Su, Weijie and Chen, Guo and Xing, Sen and Zhong, Muyan and Zhang, Qinglong and Zhu, Xizhou and Lu, Lewei and Li, Bin and Luo, Ping and Lu, Tong and Qiao, Yu and Dai, Jifeng},
  journal={arXiv preprint arXiv:2312.14238},
  year={2023}
}

@inproceedings{gsva,
  author       = {Zhuofan Xia and
                  Dongchen Han and
                  Yizeng Han and
                  Xuran Pan and
                  Shiji Song and
                  Gao Huang},
  title        = {{GSVA:} Generalized Segmentation via Multimodal Large Language Models},
  booktitle    = CVPR,
  year         = {2024},
  pages={3858--3869}
}

@article{segment-anything,
  title={Segment Anything},
  author={Kirillov, Alexander and Mintun, Eric and Ravi, Nikhila and Mao, Hanzi and Rolland, Chloe and Gustafson, Laura and Xiao, Tete and Whitehead, Spencer and Berg, Alexander C. and Lo, Wan-Yen and Doll{\'a}r, Piotr and Girshick, Ross},
  journal={arXiv preprint arXiv:2304.02643},
  year={2023}
}

@inproceedings{Hu,
  author       = {Ronghang Hu and
                  Marcus Rohrbach and
                  Trevor Darrell},
  title        = {Segmentation from Natural Language Expressions},
  booktitle    = ECCV,
  year         = {2016},
  pages={108--124}
}

@inproceedings{mattnet,
  author       = {Licheng Yu and
                  Zhe Lin and
                  Xiaohui Shen and
                  Jimei Yang and
                  Xin Lu and
                  Mohit Bansal and
                  Tamara L. Berg},
  title        = {MAttNet: Modular Attention Network for Referring Expression Comprehension},
  booktitle    = CVPR,
  year         = {2018},
  pages={1307--1315}
}

@inproceedings{vlt,
  author       = {Henghui Ding and
                  Chang Liu and
                  Suchen Wang and
                  Xudong Jiang},
  title        = {Vision-Language Transformer and Query Generation for Referring Segmentation},
  booktitle    = ICCV,
  year         = {2021},
  pages={16301--16310}
}

@inproceedings{a2d-sentence,
  author       = {Kirill Gavrilyuk and
                  Amir Ghodrati and
                  Zhenyang Li and
                  Cees G. M. Snoek},
  title        = {Actor and Action Video Segmentation From a Sentence},
  booktitle    = CVPR,
  year         = {2018},
  pages={5958--5966}
}

@inproceedings{ref-dacvis17,
  author       = {Anna Khoreva and
                  Anna Rohrbach and
                  Bernt Schiele},
  title        = {Video Object Segmentation with Language Referring Expressions},
  booktitle    = {{ACCV}},
  year         = {2018},
  pages={123-141}
}

@inproceedings{refer-ytb-vos,
  author       = {Seonguk Seo and
                  Joon{-}Young Lee and
                  Bohyung Han},
  title        = {{URVOS:} Unified Referring Video Object Segmentation Network with a Large-Scale Benchmark},
  booktitle    = ECCV,
  year         = {2020},
  pages={208--223}
}

@inproceedings{referformer,
  author       = {Jiannan Wu and
                  Yi Jiang and
                  Peize Sun and
                  Zehuan Yuan and
                  Ping Luo},
  title        = {Language as Queries for Referring Video Object Segmentation},
  booktitle    = CVPR,
  year         = {2022},
  pages={4964--4974}
}

@inproceedings{mose,
  author       = {Henghui Ding and
                  Chang Liu and
                  Shuting He and
                  Xudong Jiang and
                  Philip H. S. Torr and
                  Song Bai},
  title        = {{MOSE:} {A} New Dataset for Video Object Segmentation in Complex Scenes},
  booktitle    = ICCV,
  year         = {2023},
  pages={20167--20177}
}

@inproceedings{jhmdb,
  author       = {Hueihan Jhuang and
                  Juergen Gall and
                  Silvia Zuffi and
                  Cordelia Schmid and
                  Michael J. Black},
  title        = {Towards Understanding Action Recognition},
  booktitle    = ICCV,
  year         = {2013},
  pages={3192--3199}
}

@article{davis17,
  author       = {Jordi Pont{-}Tuset and
                  Federico Perazzi and
                  Sergi Caelles and
                  Pablo Arbel{\'{a}}ez and
                  Alexander Sorkine{-}Hornung and
                  Luc Van Gool},
  title        = {The 2017 {DAVIS} Challenge on Video Object Segmentation},
  journal      = {arXiv preprint arXiv:1704.00675},
  year         = {2017}
}

@article{internvit,
  author       = {Zhe Chen and
                  Weiyun Wang and
                  Hao Tian and
                  Shenglong Ye and
                  Zhangwei Gao and
                  Erfei Cui and
                  Wenwen Tong and
                  Kongzhi Hu and
                  et al},
  title        = {How Far Are We to GPT-4V? Closing the Gap to Commercial Multimodal Models with Open-Source Suites},
  journal      = {arXiv preprint arXiv:2404.16821},
  year         = {2024}
}

@article{qwen2,
      title={Qwen2 Technical Report}, 
      author={An Yang and Baosong Yang and Binyuan Hui and Bo Zheng and et al},
      journal={arXiv preprint arXiv:2407.10671},
      year={2024}
}

@article{qwen-vl,
  author       = {Jinze Bai and
                  Shuai Bai and
                  Shusheng Yang and
                  Shijie Wang and
                  Sinan Tan and
                  Peng Wang and
                  Junyang Lin and
                  Chang Zhou and
                  Jingren Zhou},
  title        = {Qwen-VL: {A} Frontier Large Vision-Language Model with Versatile Abilities},
  journal      = {arXiv preprint arXiv:2308.12966},
  year         = {2023}
}

@inproceedings{llava1.5,
  author       = {Haotian Liu and
                  Chunyuan Li and
                  Yuheng Li and
                  Yong Jae Lee},
  title        = {Improved Baselines with Visual Instruction Tuning},
  booktitle    = CVPR,
  year         = {2024},
  pages={26286--26296}
}

@inproceedings{bliva,
  author       = {Wenbo Hu and
                  Yifan Xu and
                  Yi Li and
                  Weiyue Li and
                  Zeyuan Chen and
                  Zhuowen Tu},
  title        = {{BLIVA:} {A} Simple Multimodal {LLM} for Better Handling of Text-Rich Visual Questions},
  booktitle    =AAAI,
  year         = {2024},
  pages={2256--2264}
}

@article{liu2022instance,
  title={Instance-Specific Feature Propagation for Referring Segmentation},
  author={Liu, Chang and Jiang, Xudong and Ding, Henghui},
  journal=TMM,
  year={2022},
  xxxxpublisher={IEEE},
  volume       = {25},
  pages        = {3657--3667}
}

@inproceedings{adamw,
  author       = {Ilya Loshchilov and
                  Frank Hutter},
  title        = {Decoupled Weight Decay Regularization},
  booktitle    = ICLR,
  year         = {2019}
}

@inproceedings{gumble-softmax,
  author       = {Eric Jang and
                  Shixiang Gu and
                  Ben Poole},
  title        = {Categorical Reparameterization with Gumbel-Softmax},
  booktitle    = ICLR,
  year         = {2017}
}

@inproceedings{mask2former,
  author       = {Bowen Cheng and
                  Ishan Misra and
                  Alexander G. Schwing and
                  Alexander Kirillov and
                  Rohit Girdhar},
  title        = {Masked-attention Mask Transformer for Universal Image Segmentation},
  booktitle    = CVPR,
  year         = {2022},
  pages={1280--1289}
}

@inproceedings{trick,
  author       = {A{\"{a}}ron van den Oord and
                  Oriol Vinyals and
                  Koray Kavukcuoglu},
  title        = {Neural Discrete Representation Learning},
  booktitle    = NeurIPS,
  year         = {2017}
}

@inproceedings{dice,
  author       = {Fausto Milletari and
                  Nassir Navab and
                  Seyed{-}Ahmad Ahmadi},
  title        = {V-Net: Fully Convolutional Neural Networks for Volumetric Medical Image Segmentation},
  booktitle    = {3DV},
  year         = {2016}
}

@inproceedings{lbdt,
  author       = {Zihan Ding and
                  Tianrui Hui and
                  Junshi Huang and
                  Xiaoming Wei and
                  Jizhong Han and
                  Si Liu},
  title        = {Language-Bridged Spatial-Temporal Interaction for Referring Video Object Segmentation},
  booktitle    = CVPR,
  year         = {2022},
  pages={4954--4963}
}

@inproceedings{mlsa,
  author       = {Dongming Wu and
                  Xingping Dong and
                  Ling Shao and
                  Jianbing Shen},
  title        = {Multi-Level Representation Learning with Semantic Alignment for Referring Video Object Segmentation},
  booktitle    = CVPR,
  year         = {2022},
  pages={4986--4995}
}

@inproceedings{mttr,
  author       = {Dongming Wu and
                  Xingping Dong and
                  Ling Shao and
                  Jianbing Shen},
  title        = {Multi-Level Representation Learning with Semantic Alignment for Referring Video Object Segmentation},
  booktitle    = CVPR,
  year         = {2022},
  pages={4986--4995}
}

@inproceedings{html,
  author       = {Mingfei Han and
                  Yali Wang and
                  Zhihui Li and
                  Lina Yao and
                  Xiaojun Chang and
                  Yu Qiao},
  title        = {{HTML:} Hybrid Temporal-scale Multimodal Learning Framework for Referring Video Object Segmentation},
  booktitle    = ICCV,
  year         = {2023},
  pages={13368--13377}
}

@inproceedings{r2vos,
  author       = {Xiang Li and
                  Jinglu Wang and
                  Xiaohao Xu and
                  Xiao Li and
                  Bhiksha Raj and
                  Yan Lu},
  title        = {Robust Referring Video Object Segmentation with Cyclic Structural Consensus},
  booktitle    = {ICCV},
  year         = {2023},
  pages={22179--22188}
}

@inproceedings{sgmg,
  author       = {Bo Miao and
                  Mohammed Bennamoun and
                  Yongsheng Gao and
                  Ajmal Mian},
  title        = {Spectrum-guided Multi-granularity Referring Video Object Segmentation},
  booktitle    = ICCV,
  year         = {2023},
  pages={920--930}
}

@inproceedings{onlinerefer,
  author       = {Dongming Wu and
                  Tiancai Wang and
                  Yuang Zhang and
                  Xiangyu Zhang and
                  Jianbing Shen},
  title        = {OnlineRefer: {A} Simple Online Baseline for Referring Video Object Segmentation},
  booktitle    = ICCV,
  year         = {2023},
  pages={2749--2758}
}

@inproceedings{tempcd,
  author       = {Jiajin Tang and
                  Ge Zheng and
                  Sibei Yang},
  title        = {Temporal Collection and Distribution for Referring Video Object Segmentation},
  booktitle    = ICCV,
  year         = {2023},
  pages={15420--15430}
}

@inproceedings{soc,
  author       = {Zhuoyan Luo and
                  Yicheng Xiao and
                  Yong Liu and
                  Shuyan Li and
                  Yitong Wang and
                  Yansong Tang and
                  Xiu Li and
                  Yujiu Yang},
  title        = {{SOC:} Semantic-Assisted Object Cluster for Referring Video Object Segmentation},
  booktitle    = NeurIPS,
  year         = {2023},
  pages={26425--26437}
}

@article{vlt+tc,
  author       = {Henghui Ding and
                  Chang Liu and
                  Suchen Wang and
                  Xudong Jiang},
  title        = {{VLT:} Vision-Language Transformer and Query Generation for Referring Segmentation},
  journal      = PAMI,
  year         = {2023},
  volume       ={45},
  number       ={6},
  pages        ={7900-7916}
}

@article{trackgpt,
  author       = {Jiawen Zhu and
                  Zhi{-}Qi Cheng and
                  Jun{-}Yan He and
                  Chenyang Li and
                  Bin Luo and
                  Huchuan Lu and
                  Yifeng Geng and
                  Xuansong Xie},
  title        = {Tracking with Human-Intent Reasoning},
  journal      = {arXiv preprint arXiv:2312.17448},
  year         = {2023}
}

@inproceedings{xmem,
  author       = {Ho Kei Cheng and
                  Alexander G. Schwing},
  title        = {XMem: Long-Term Video Object Segmentation with an Atkinson-Shiffrin Memory Model},
  booktitle    = ECCV,
  year         = {2022},
  pages={640--658}
}

@inproceedings{chatunivi,
  author       = {Peng Jin and
                  Ryuichi Takanobu and
                  Wancai Zhang and
                  Xiaochun Cao and
                  Li Yuan},
  title        = {Chat-UniVi: Unified Visual Representation Empowers Large Language Models with Image and Video Understanding},
  booktitle    = CVPR,
  year         = {2024},
  pages={13700--13710}
}

@inproceedings{imagepoints,
  author       = {Xu Ma and
                  Yuqian Zhou and
                  Huan Wang and
                  Can Qin and
                  Bin Sun and
                  Chang Liu and
                  Yun Fu},
  title        = {Image as Set of Points},
  booktitle    = ICLR,
  year         = {2023}
}

@inproceedings{dynamic-vit,
  author       = {Yongming Rao and
                  Wenliang Zhao and
                  Benlin Liu and
                  Jiwen Lu and
                  Jie Zhou and
                  Cho{-}Jui Hsieh},
  title        = {DynamicViT: Efficient Vision Transformers with Dynamic Token Sparsification},
  booktitle    = NeurIPS,
  year         = {2021},
  pages={13937--13949}
}

@inproceedings{testa,
  author       = {Shuhuai Ren and
                  Sishuo Chen and
                  Shicheng Li and
                  Xu Sun and
                  Lu Hou},
  title        = {{TESTA:} Temporal-Spatial Token Aggregation for Long-form Video-Language Understanding},
  booktitle    = {{EMNLP}},
  year         = {2023},
  pages={932-947}
}

@article{dualfocus,
  author       = {Yuhang Cao and
                  Pan Zhang and
                  Xiaoyi Dong and
                  Dahua Lin and
                  Jiaqi Wang},
  title        = {DualFocus: Integrating Macro and Micro Perspectives in Multi-modal Large Language Models},
  journal      = {arXiv preprint arXiv:2402.14767},
  year         = {2024}
}

@inproceedings{groupvit,
  author       = {Jiarui Xu and
                  Shalini De Mello and
                  Sifei Liu and
                  Wonmin Byeon and
                  Thomas M. Breuel and
                  Jan Kautz and
                  Xiaolong Wang},
  title        = {GroupViT: Semantic Segmentation Emerges from Text Supervision},
  booktitle    = CVPR,
  year         = {2022},
  pages={18113--18123}
}

@inproceedings{losh,
  author       = {Linfeng Yuan and
                  Miaojing Shi and
                  Zijie Yue and
                  Qijun Chen},
  title        = {LoSh: Long-Short Text Joint Prediction Network for Referring Video Object Segmentation},
  booktitle    = CVPR,
  year         = {2024},
  pages={14001--14010}
}

@inproceedings{transformer,
  author       = {Ashish Vaswani and
                  Noam Shazeer and
                  Niki Parmar and
                  Jakob Uszkoreit and
                  Llion Jones and
                  Aidan N. Gomez and
                  Lukasz Kaiser and
                  Illia Polosukhin},
  title        = {Attention is All you Need},
  booktitle    = NeurIPS,
  year         = {2017},
}

@article{roberta,
  author       = {Yinhan Liu and
                  Myle Ott and
                  Naman Goyal and
                  Jingfei Du and
                  Mandar Joshi and
                  Danqi Chen and
                  Omer Levy and
                  Mike Lewis and
                  Luke Zettlemoyer and
                  Veselin Stoyanov},
  title        = {RoBERTa: {A} Robustly Optimized {BERT} Pretraining Approach},
  journal      = {arXiv preprint arXiv:1907.11692},
  year         = {2019}
}

@inproceedings{x-decoder,
  author       = {Xueyan Zou and
                  Zi{-}Yi Dou and
                  Jianwei Yang and
                  Zhe Gan and
                  Linjie Li and
                  Chunyuan Li and
                  Xiyang Dai and
                  Harkirat Behl and
                  Jianfeng Wang and
                  Lu Yuan and
                  Nanyun Peng and
                  Lijuan Wang and
                  Yong Jae Lee and
                  Jianfeng Gao},
  title        = {Generalized Decoding for Pixel, Image, and Language},
  booktitle    = CVPR,
  year         = {2023},
  pages={15116--15127}
}

@inproceedings{SEEM,
  author       = {Xueyan Zou and
                  Jianwei Yang and
                  Hao Zhang and
                  Feng Li and
                  Linjie Li and
                  Jianfeng Wang and
                  Lijuan Wang and
                  Jianfeng Gao and
                  Yong Jae Lee},
  title        = {Segment Everything Everywhere All at Once},
  booktitle    = NeurIPS,
  year         = {2023},
  pages={19769--19782}
}

@article{grounded-sam,
  author       = {Tianhe Ren and
                  Shilong Liu and
                  Ailing Zeng and
                  Jing Lin and
                  Kunchang Li and
                  He Cao and
                  Jiayu Chen and
                  Xinyu Huang and
                  Yukang Chen and
                  Feng Yan and
                  Zhaoyang Zeng and
                  Hao Zhang and
                  Feng Li and
                  Jie Yang and
                  Hongyang Li and
                  Qing Jiang and
                  Lei Zhang},
  title        = {Grounded {SAM:} Assembling Open-World Models for Diverse Visual Tasks},
  journal={arXiv preprint arXiv:2401.14159},
  year         = {2024}
}

@inproceedings{ovseg,
  author       = {Feng Liang and
                  Bichen Wu and
                  Xiaoliang Dai and
                  Kunpeng Li and
                  Yinan Zhao and
                  Hang Zhang and
                  Peizhao Zhang and
                  Peter Vajda and
                  Diana Marculescu},
  title        = {Open-Vocabulary Semantic Segmentation with Mask-adapted {CLIP}},
  booktitle    = CVPR,
  year         = {2023},
  pages={7061--7070}
}

@inproceedings{mcn,
  author       = {Gen Luo and
                  Yiyi Zhou and
                  Xiaoshuai Sun and
                  Liujuan Cao and
                  Chenglin Wu and
                  Cheng Deng and
                  Rongrong Ji},
  title        = {Multi-Task Collaborative Network for Joint Referring Expression Comprehension and Segmentation},
  booktitle    = CVPR,
  year         = {2020},
  pages={10031--10040}
}

@inproceedings{cris,
  author       = {Zhaoqing Wang and
                  Yu Lu and
                  Qiang Li and
                  Xunqiang Tao and
                  Yandong Guo and
                  Mingming Gong and
                  Tongliang Liu},
  title        = {{CRIS:} CLIP-Driven Referring Image Segmentation},
  booktitle    = CVPR,
  year         = {2022},
  pages={11676--11685}
}

@inproceedings{refcoco+,
  author       = {Sahar Kazemzadeh and
                  Vicente Ordonez and
                  Mark Matten and
                  Tamara L. Berg},
  title        = {ReferItGame: Referring to Objects in Photographs of Natural Scenes},
  booktitle    = {{EMNLP}},
  year         = {2014},
  pages={787--798}
}

@inproceedings{refcocog,
  author       = {Junhua Mao and
                  Jonathan Huang and
                  Alexander Toshev and
                  Oana Camburu and
                  Alan L. Yuille and
                  Kevin Murphy},
  title        = {Generation and Comprehension of Unambiguous Object Descriptions},
  booktitle    = CVPR,
  year         = {2016},
  pages={11--20}
}

@inproceedings{glus,
  title={Glus: Global-local reasoning unified into a single large language model for video segmentation},
  author={Lin, Lang and Yu, Xueyang and Pang, Ziqi and Wang, Yu-Xiong},
  booktitle={CVPR},
  year={2025},
  pages={8658--8667}
}

@inproceedings{devil,
  title={The devil is in temporal token: High quality video reasoning segmentation},
  author={Gong, Sitong and Zhuge, Yunzhi and Zhang, Lu and Yang, Zongxin and Zhang, Pingping and Lu, Huchuan},
  booktitle={CVPR},
  year={2025},
  pages={29183--29192}
}

@article{Veason-R1,
  title={Reinforcing video reasoning segmentation to think before it segments},
  author={Gong, Sitong and Zhang, Lu and Zhuge, Yunzhi and Jia, Xu and Zhang, Pingping and Lu, Huchuan},
  journal={arXiv preprint arXiv:2508.11538},
  year={2025}
}

@article{grpo,
  title={Deepseekmath: Pushing the limits of mathematical reasoning in open language models},
  author={Shao, Zhihong and Wang, Peiyi and Zhu, Qihao and Xu, Runxin and Song, Junxiao and Bi, Xiao and Zhang, Haowei and Zhang, Mingchuan and Li, YK and Wu, Yang and others},
  journal={arXiv preprint arXiv:2402.03300},
  year={2024}
}

\begin{IEEEbiography}[{\includegraphics[width=1in,height=1.25in,clip,keepaspectratio]{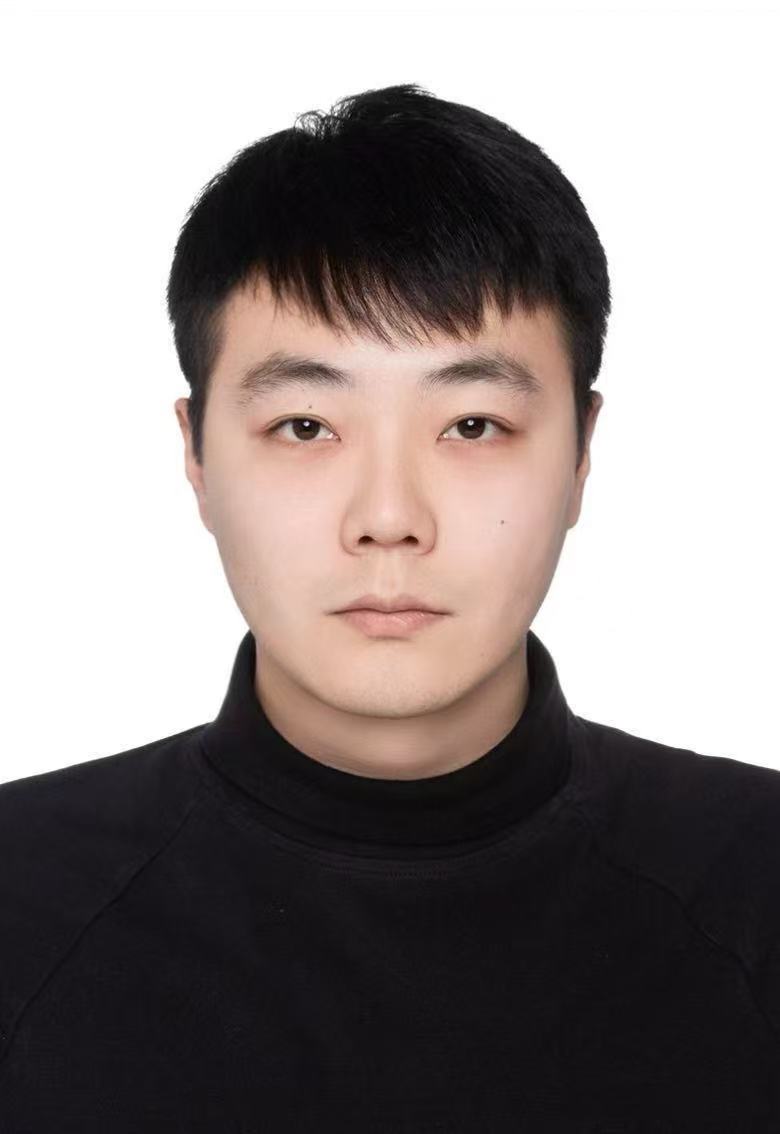}}]{Jinyu Liu} received the M.S. degree from Fudan University, Shanghai, China, in 2023. He is currently a Ph.D. student at College of Computer Science and Artificial Intelligence, Fudan University, Shanghai, China. Prior to that, he was a Research Intern at Tencent. His research interests include computer vision and multi-modal learning.
\end{IEEEbiography}

\begin{IEEEbiography}[{\includegraphics[width=1in,height=1.25in,clip,keepaspectratio]{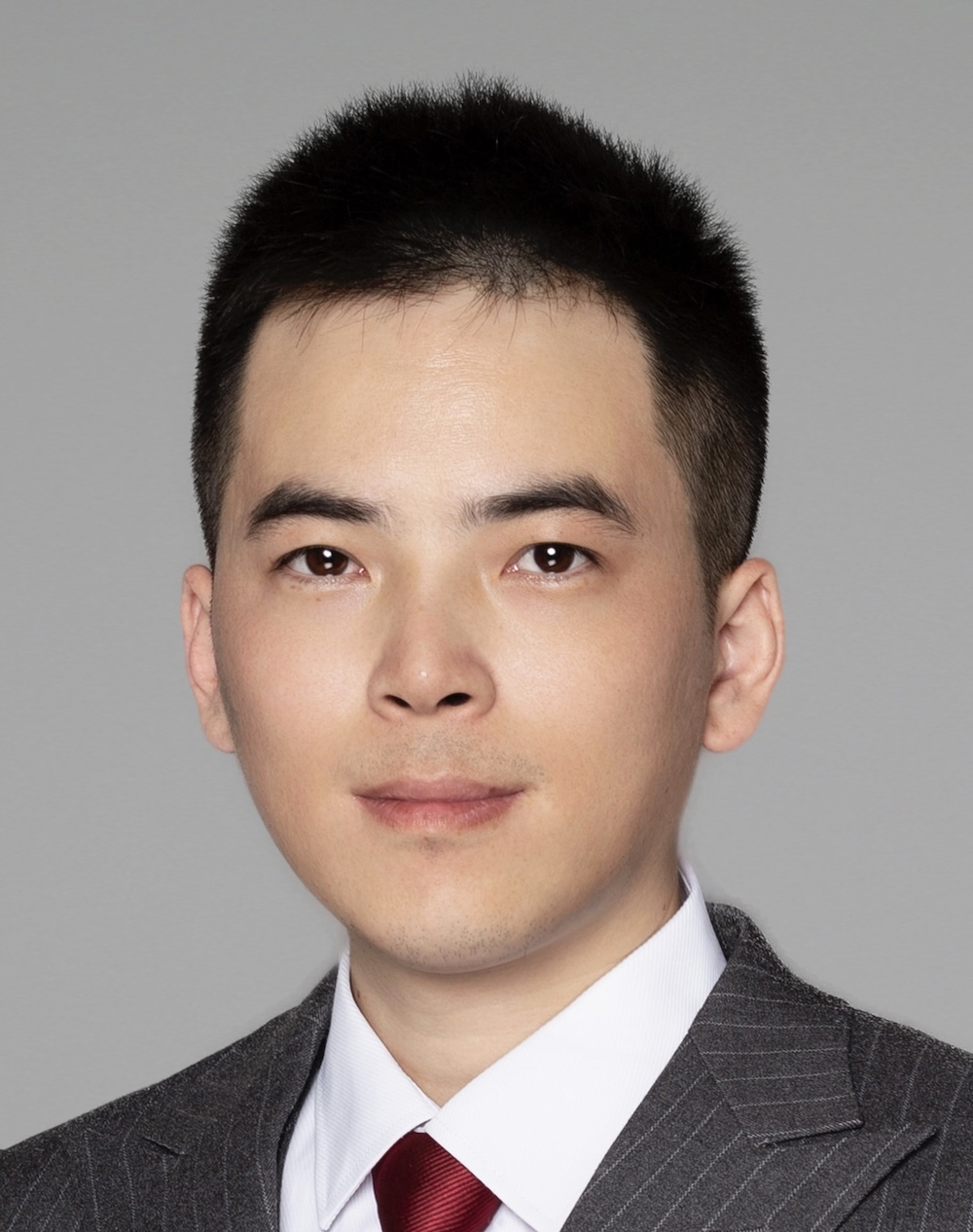}}]{Henghui Ding} (Member, IEEE) received the B.E. degree from Xi'an Jiaotong University, China, in 2016, and the Ph.D. degree from Nanyang Technological University (NTU), Singapore, in 2020. He was a Research Scientist at ByteDance, a Postdoctoral Researcher at ETH Zurich and NTU. He is currently a Professor at Fudan University, Shanghai, China. He serves as an Associate Editor for IEEE Transactions on Image Processing (TIP) and Pattern Recognition (PR), and regularly serves/served as a Senior Area Chair or Area Chair of top conferences such as CVPR, NeurIPS, ICLR, ICML, ECCV, AAAI, and ACM MM. His research interests include computer vision and machine learning. 
\end{IEEEbiography}

\begin{IEEEbiography}[{\includegraphics[width=1in,height=1.25in,clip,keepaspectratio]{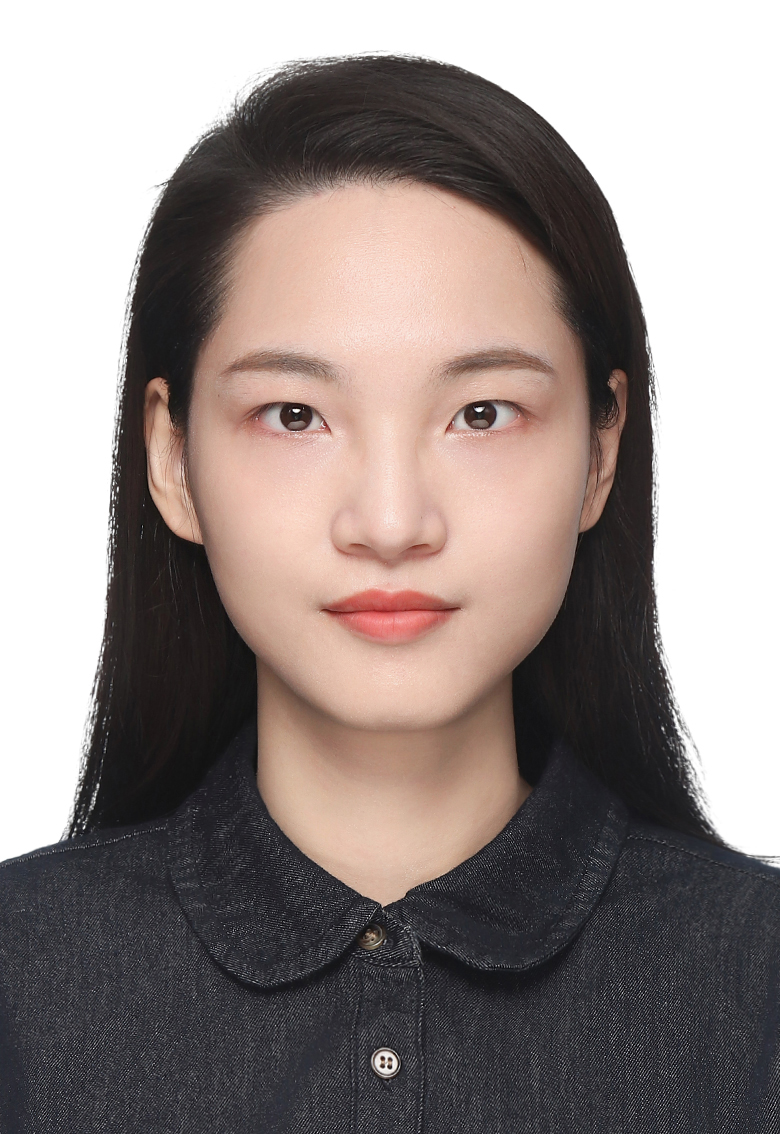}}]{Shuting He} received the B.E. degree from Xiamen University (XMU), Xiamen, China, in 2018, and the Ph.D. degree from Zhejiang University (ZJU), Hangzhou, China, in 2023. She is currently a tenure-track Assistant Professor with Shanghai University of Finance and Economics (SUFE). Prior to that, she was a Research Fellow with Nanyang Technological University (NTU), Singapore. She serves as an Area Chair for CVPR, NeurIPS, ICLR, and BMVC. Her research interests include computer vision and machine learning.  
\end{IEEEbiography}

\begin{IEEEbiography}[{\includegraphics[width=1in,height=1.25in,clip,keepaspectratio]{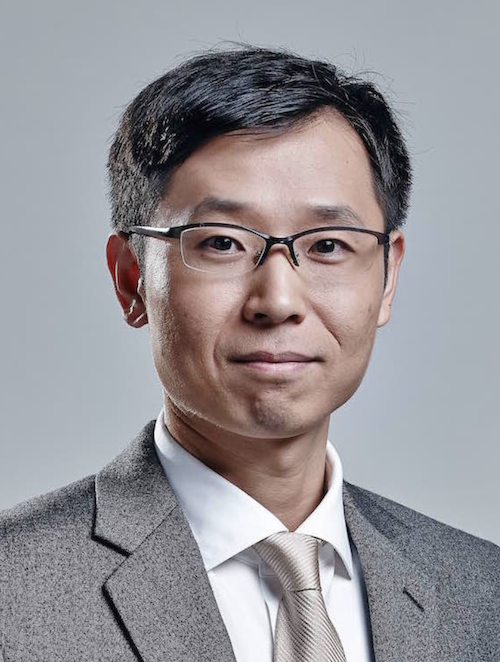}}]{Yu-Gang Jiang} (Fellow, IEEE) received the PhD degree in Computer Science from City University of Hong Kong in 2009 and worked as a Postdoctoral Research Scientist at Columbia University, New York, during 2009-2011. He is currently a Distinguished Professor of Computer Science at Fudan University, Shanghai, China. His research lies in the areas of multimedia, computer vision, embodied AI and trustworthy AI. His research has led to the development of innovative AI tools that have been used in many practical applications like defect detection for high-speed railway infrastructures. His open-source video analysis toolkits and datasets such as CU-VIREO374, CCV, THUMOS, FCVID and WildDeepfake have been widely used in both academia and industry. He currently serves as Chair of ACM Shanghai Chapter and Associate Editor of several international journals. For contributions to large-scale and trustworthy video analysis, he was elected to Fellow of IEEE, IAPR, and CCF.
\end{IEEEbiography}

\end{document}